\documentclass[natbib,twocolumn]{svjour3}
\smartqed

\usepackage{mathptmx}
\usepackage[colorlinks,citecolor=black,urlcolor=black]{hyperref}
\usepackage{amsmath,amssymb}
\usepackage{tensor}
\usepackage{graphicx}
\usepackage{caption}
\usepackage{subcaption}
\usepackage{overpic}
\usepackage{rotating}
\usepackage{verbatim}
\graphicspath{{./fig/}}
\usepackage{algorithm}
\usepackage{algorithmic}
\usepackage{color}
\usepackage{soul}
\usepackage{bm}
\usepackage{multirow}
\usepackage{multicol}
\usepackage{arydshln}
\usepackage{xspace}
\usepackage{graphicx}
\usepackage{todonotes}
\usepackage{flushend}
\usepackage{float}

\DeclareCaptionSubType*[Alph]{table}
\DeclareCaptionLabelFormat{mystyle}{Table~\bothIfFirst{#1}{ ̃}#2}
\DeclareMathOperator*{\argmin}{arg\!\min}
\DeclareMathOperator*{\argmax}{arg\!\max}

\newcommand\ol{OL\xspace}
\newcommand\slap{SLAP\xspace}
\newcommand\algo{STEAP\xspace}

\begin{document}

\title{\bf STEAP: simultaneous trajectory estimation and planning
}

\titlerunning{STEAP}

\author{
	Mustafa Mukadam\textsuperscript{1}\textsuperscript{*}
	\and
	Jing Dong\textsuperscript{1}\textsuperscript{*}
	\and
	Frank Dellaert\textsuperscript{1}
	\and
	Byron Boots\textsuperscript{1}
}

\authorrunning{Mukadam et al.}

\institute{
	Mustafa Mukadam \\
	{mmukadam3@gatech.edu} \\
	\\
	Jing Dong \\
	{jdong@gatech.edu} \\
	\\
	Frank Dellaert \\
	{frank@cc.gatech.edu} \\
	\\
	Byron Boots \\
	{bboots@cc.gatech.edu} \\
	\\
	\textsuperscript{*}Equal contribution to this article \\
	\textsuperscript{1}Institute for Robotics \& Intelligent Machines,
	Georgia Institute of Technology,
	Atlanta, GA, USA
}

\maketitle


\begin{abstract}
We present a unified probabilistic framework for simultaneous trajectory estimation and planning (STEAP). Estimation and planning problems are usually considered separately, however, within our framework we show that solving them simultaneously can be more accurate and efficient. The key idea is to compute the full continuous-time trajectory from start to goal at each time-step.  While the robot traverses the trajectory, the history portion of the trajectory signifies the solution to the estimation problem, and the future portion of the trajectory signifies a solution to the planning problem. Building on recent probabilistic inference approaches to continuous-time localization and mapping and continuous-time motion planning, we solve the joint problem by iteratively recomputing the \emph{maximum a posteriori} trajectory conditioned on all available sensor data and cost information. Our approach can contend with high-degree-of-freedom (DOF) trajectory spaces, uncertainty due to limited sensing capabilities, model inaccuracy, the stochastic effect of executing actions, and can find a solution in real-time. We evaluate our framework empirically in both simulation and on a mobile manipulator.
\keywords{Estimation \and Motion planning \and Replanning \and Trajectory optimization \and Probabilistic inference \and Factor graphs \and Gaussian processes}
\end{abstract}


\section{Introduction}\label{sec:intro}

Trajectory estimation and planning are both important capabilities for autonomous robot navigation. Trajectory estimation is fundamentally backward-looking: the robot estimates a trajectory of previous states that are consistent with a history of noisy and incomplete sensor data. Conversely, planning is fundamentally forward looking: starting from an estimate of its current state, the robot optimizes a trajectory of future states to minimize a cost function and achieve a feasible solution.

In this work, we provide a unified approach to trajectory estimation and planning. Our key insight is that both these problems are inherently variants of trajectory optimization and can therefore be combined to remove the redundancy present in a traditional two step process. The idea is to compute the complete continuous-time trajectory from start to goal at each time-step, such that given the current time-step, the solutions to the estimation problem (history of the trajectory) and the planning problem (future of the trajectory) automatically fall out. Additionally, performing this joint optimization allows information to flow between estimation and planning resulting in mutual benefits. This problem can be quite difficult to solve; the robot must contend with a potentially high-degree-of-freedom (DOF) trajectory space, uncertainty due to limited sensing capabilities, model inaccuracy, and the stochastic effect of executing actions. For the solution to be practical, it must be generated in (faster than) real-time. 

\begin{figure}[!t]
\centering
\includegraphics[width=1\linewidth]{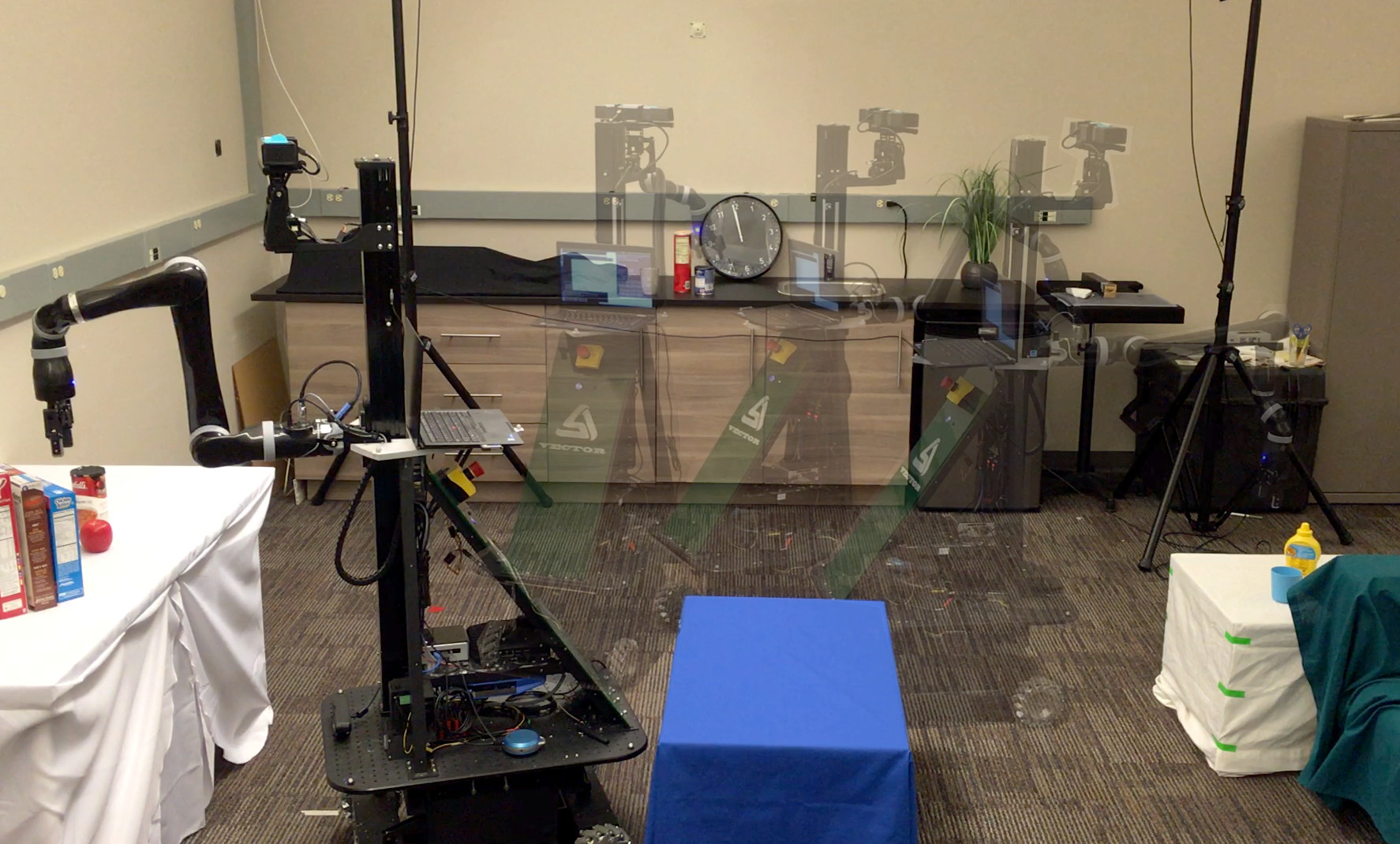}  
\caption{The Vector mobile manipulator, with an omni-drive base and a 6-DOF Kinova JACO2 arm, is solving the \algo problem. The task involves picking up an object from the white table on the right and dropping it off on the white table on the left. The semi-transparent robots show the trajectory taken, while the solid robot is the goal configuration.}
\label{fig:vector}
\end{figure}

We propose a solution to the problem of simultaneous trajectory estimation and planning (STEAP) by viewing trajectory optimization as probabilistic inference and building on recent approaches to continuous-time localization and mapping \citep{anderson2015batch,Yan17ras} and continuous-time motion planning \citep{Dong-RSS-16,mukadam2017continuous}. We represent the continuous-time trajectory as a function mapping time to robot states and model the trajectory distribution as a Gaussian process (GP) \citep{barfoot2014batch,Mukadam-ICRA-16}. At each time-step, we incrementally (re)estimate the entire continuous-time trajectory, as new sensor data or cost information is encountered, by iteratively recomputing the \emph{maximum a posteriori} (MAP) trajectory conditioned on all the available sensor data and cost information. In general, sensor data can include various measurements from proprioception (encoders, inertial measurement unit (IMU), etc) or perception for external information (cameras, LIDARs, etc). On the other hand, cost information can include full or partial information know before (start, goal, map of the environment, etc) or constraints and information encountered during execution (changing goal, orientation constraints on the end effector for a portion of the trajectory when for example a cup with liqiud is held, etc).

We formulate the STEAP problem on a single probabilistic graphical model and seek the MAP function with incremental inference \citep{kaess2011isam2}. This allows us to exploit the underlying sparsity of the problem and avoid re-solving it from scratch as new information is encountered. In our approach the trajectory is only updated where required, dramatically reducing the overall computational burden and enabling a faster-than-real-time solution. We also provide theoretical insight on the connections between our approach and various methods in mapping, estimation, and planning in the context of solving them as inference on graphical models. To better accommodate mobile manipulation problems, we build on recent work by \citet{anderson2015full} on continuous-time trajectory estimation on SE(3) and extend \citet{Dong-RSS-16} to plan trajectories on Lie groups. We implement our framework for solving STEAP and perform several experiments to evaluate our approach in simulation and on the Vector mobile manipulator (Fig.~\ref{fig:vector}), and show that our framework is able to incrementally integrate real-world sensor data and directly update its trajectory estimate and motion plan in real-time.
This paper is an extended and revised version of our conference paper \citep{Mukadam-RSS-17}. In particular,
\begin{itemize}
\item We summarize the incremental inference via Bayes trees approach \citep{kaess2011isam2} in Section~\ref{sec:incinf}, specifically in the context of solving the STEAP problem.
\item We provide a detailed explanation on formulating sparse GPs on Lie groups \citep{dong2017arxiv} in Section~\ref{sec:gp_prior}.
\item We present updated experiments on a harder dataset for the planar robot and new experiments with a 18-DOF PR2 robot in simulation.
\item We provide further insight in to our approach by adding a discussion on limitations and future work in Section~\ref{sec:diss}.
\end{itemize}


\section{Related Work}\label{sec:rel_work}

By viewing trajectory estimation and motion planning as inference, we are able to borrow and combine tools from different areas of robotics. The Simultaneous Localization and Mapping (SLAM) community has focused on efficient optimization algorithms for many years. One of the more successful approaches is the Smoothing and Mapping (SAM) family of algorithms \citep{dellaert2006square} that formulates SLAM as inference on a factor graph \citep{kschischang2001factor} and exploits the sparsity of the underlying large-scale linear systems to perform inference efficiently. Given new sensor data, incremental Smoothing and Mapping (iSAM) \citep{kaess2008isam, kaess2011isam2} exploits the structure of the problem to efficiently update the solution rather than resolving the entire problem from scratch. Recently, \citet{tong2013gaussian} introduced a continuous-time formulation of the SAM problem, in which the robot trajectory is a function that maps any time to a robot state. The problem of estimating this function along with landmark locations has been dubbed simultaneous trajectory estimation and mapping (STEAM). This approach was further extended in \citet{barfoot2014batch} to take advantage of the sparse structure inherent in the STEAM problem, in \citet{Yan17ras} to efficiently and incrementally update the solution, and in \citet{Dong-ICRA-17} to 4D mapping problems. The resulting algorithms speed up solution time and can be viewed as continuous-time analogs of the original square-root SAM algorithm in \citet{dellaert2006square} and the iSAM2 algorithm in \citet{kaess2011isam2}.

While probabilistic inference is frequently used as a foundation for state estimation and localization, it is only recently that these techniques have been used for planning. The duality between linear estimation and control has long been established \citep{kalman1960new}, but solutions to estimation and control problems have, for the most part, evolved independently within their own subfields. In the last decade this has begun to change. The optimization-inference duality has been shown to extend to planning and optimal control \citep{todorov2008general} with some early work in this direction looking at solving Markov decision processes (MDP) \citep{attias2003planning}. Several researchers have recently proposed a probabilistic inference perspective on planning and control problems, leveraging expectation maximization \citep{toussaint2006probabilistic,levine2013variational}, expectation propagation \citep{toussaint2009robot}, KL-minimization \citep{rawlik2012stochastic}, and efficient inference on factor graphs \citep{Dong-RSS-16,mukadam2017continuous,Mukadam-ICRA-17,Huang-ICRA-17,pmlr-v78-rana17a}. Interestingly, the incremental inference technique \citep{kaess2011bayes} used in \citet{Dong-RSS-16} to solve replanning problems is the same as originally used in \citet{kaess2011isam2} to solve SLAM problems. We exploit this idea to solve our more general class of simultaneous trajectory estimation and planning problems.

Efficient replanning algorithms for navigation are an active area of research \citep{koenig2005fast,ferguson2006replanning}, but most previous work is difficult to extend to real, high-dimensional systems, is computationally expensive, or does not incorporate uncertainty in the robot's state estimate. Recent work in simultaneous localization and planning (SLAP) attempts to unify robot localization and planning, with early work using HMMs \citep{penny2014simultaneous}, more recent approaches designed for dynamic environments \citep{agha2015simultaneous,rafieisakhaei2016non}, and new approaches \citep{ta2014factor} that combine state estimation and model predictive control (MPC) \citep{camacho2013model}. Unfortunately, these approaches are too computationally expensive due to the MPC style re-evaluation of the new plan, which is compounded with high DOF systems in cluttered environments. In this work, we tackle the simultaneous trajectory estimation and planning (STEAP) problem within a unified probabilistic inference framework. The STEAP problem can be considered as a generalization of the SLAP problem in that the goal of STEAP is to compute the full continuous-time trajectory conditioned on observations and costs in both the past and the future. By contrast, SLAP only computes the current state estimate and the new plan.


\section{Background: Trajectory Optimization as Probabilistic Inference}\label{sec:inf}

Following previous work on both STEAM problems \citep{barfoot2014batch,Yan17ras} and Gaussian process motion planning \citep{Dong-RSS-16,mukadam2017continuous}, we view the problem of estimating or optimizing continuous-time trajectories as probabilistic inference. We represent the trajectory as a continuous-valued function mapping time $t$ to robot states ${\bm \theta}(t)$. The goal is to find the \emph{maximum a posteriori} (MAP) continuous-time trajectory given a prior distribution on the space of trajectories and a likelihood function.

\subsection{Trajectory prior}\label{sec:traj_prior}
A prior distribution over trajectories can be defined as a vector-valued Gaussian process $\bm{\theta}(t) \sim {GP}(\bm{\mu}(t), \mathbf{{K}}(t, t'))$, where $\bm{\mu}(t)$ is a vector-valued mean function and $\mathbf{{K}}(t,t')$ is a matrix-valued covariance function. For any collection of times ${\bm t} = \{ t_0, \dots, t_N\}$, $\bm{\theta}$ has a joint Gaussian distribution
\begin{equation}
\bm{\theta} \doteq \begin{bmatrix} \bm{\theta}_0 & \hdots & \bm{\theta}_N \end{bmatrix}^\top \sim {N}(\bm{\mu}, \mathbf{{K}})
\end{equation}
with mean vector $\bm{\mu}$ and covariance kernel $\mathbf{{K}}$ defined as
\begin{equation} 
\bm{\mu} \doteq \begin{bmatrix} \bm{\mu}(t_0) & \hdots & \bm{\mu}(t_N) \end{bmatrix}^\top\hspace{-2mm}, \,
\mathbf{{K}}  \doteq  [\mathbf{{K}}(t_i, t_j)]\Bigr|_{ij, 0 \leq i,j \leq N}.
\end{equation}
The prior distribution is then defined by the GP mean $ \bm{\mu}$ and covariance $\mathbf{{K}}$
\begin{equation}\label{eq:prior}
	p(\bm{\theta}) \propto \exp \bigg\{ - \frac{1}{2} \parallel \bm{\theta} - \bm{\mu} \parallel^{2}_{\mathbf{{K}}} \bigg\}.
\end{equation}
The prior encodes information about the system that is known \emph{a priori}. For example, in robotic state estimation problems, a structured GP prior may encourage trajectories to follow known system dynamics, e.g. the robot velocity changes smoothly \citep{barfoot2014batch, tong2013gaussian}. In motion planning, the prior is selected to encourage higher-order derivatives of the system configuration to be minimized \citep{Dong-RSS-16,mukadam2017continuous}. The prior we use in our implementation is detailed in Section~\ref{sec:prior}.

\subsection{Likelihood function}\label{sec:likelihood}
The likelihood function encodes information about a particular problem instance. For example, in STEAM problems, the likelihood function encourages posterior trajectories to be consistent with proprioceptive or landmark observations \citep{barfoot2014batch}, while in motion planning problems the likelihood function encourages posterior trajectories to be collision-free \citep{Dong-RSS-16}. 

Let $\mathbf{e}$ be a collection of random binary events. Examples of events include collision, receiving a sensor measurement, or reaching a goal. The likelihood function is the conditional distribution $l(\bm{\theta} ; \mathbf{e}) = p(\mathbf{e} | \bm{\theta})$, which specifies the probability of events $\mathbf{e}$ given a trajectory $\bm{\theta}$. We define the likelihood as a distribution in the exponential family
\begin{align}
	l(\bm{\theta} ; \mathbf{e}) \propto  \exp \bigg\{ - \frac{1}{2} \parallel \bm{h}(\bm{\theta},\mathbf{e}) \parallel^{2}_{\bm{\Sigma}} \bigg\} \label{eq:likelihood}
\end{align}
where $\bm{h}(\bm{\theta},\mathbf{e})$ can be any vector-valued cost function with covariance matrix $\bm{\Sigma}$. The specific likelihood used in our implementation is detailed in Section~\ref{sec:steap_factors}.

\begin{figure*}[t]
	\centering
	\begin{subfigure}[b]{0.28\textwidth}
		\centering
		\includegraphics[width=0.9\linewidth]{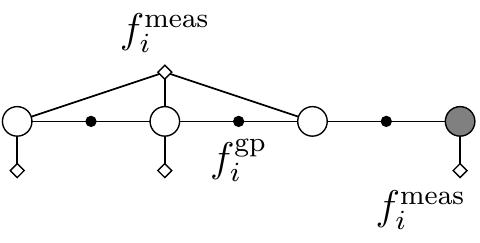}
		\caption{}
	\end{subfigure}
	\begin{subfigure}[b]{0.28\textwidth}
		\centering
		\includegraphics[width=0.9\linewidth]{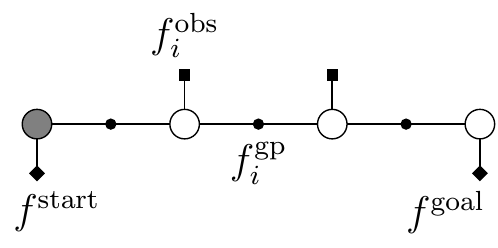}
		\caption{}
	\end{subfigure}
	\begin{subfigure}[b]{0.4\textwidth}
		\centering
		\includegraphics[width=0.99\linewidth]{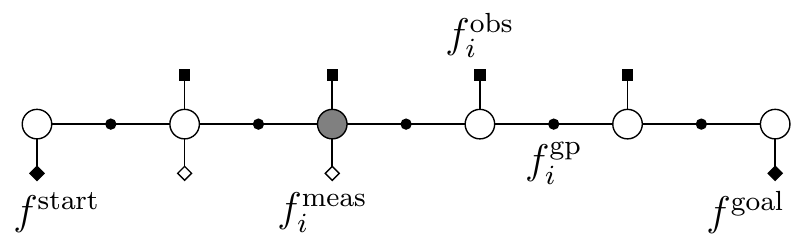}
		\caption{}
	\end{subfigure}
	\protect\caption{Example factor graph representation of (a) STEAM, (b) GPMP2, and (c) STEAP. Gray node shows current time-step.}
	\label{fig:fgs}
\end{figure*}

\subsection{Computing the MAP trajectory}\label{sec:MAP}
The posterior distribution of the trajectory given the events can be written in terms of the prior and the likelihood using the Bayes rule
\begin{equation}
	p(\bm{\theta} | \mathbf{e}) \propto p(\bm{\theta})  p( \mathbf{e} | \bm{\theta}). \label{eq:posterior}
\end{equation}
Then, we can compute the \emph{maximum a posteriori} (MAP) trajectory
\begin{align}
\bm{\theta}^*
&= \argmax_{\bm{\theta}} \big\{ p(\bm{\theta} | \mathbf{e}) \big\} = \argmax_{\bm{\theta}} \big\{ p(\bm{\theta})  p( \mathbf{e} | \bm{\theta}) \big\} \\
&= \argmin_{\bm{\theta}} \big\{- \log \big( p(\bm{\theta})
p({\bf e}| \bm{\theta}) \big) \big\} \\
&= \argmin_{\bm{\theta}} \bigg\{ \frac{1}{2} \parallel \bm{\theta} - \bm{\mu}
\parallel^{2}_{\mathbf{{K}}} + \frac{1}{2} \parallel \bm{h}(\bm{\theta},\mathbf{e})
\parallel^{2}_{\bm{\Sigma}}  \bigg\} \label{eq:least_square1}
\end{align}
where Eq.~\eqref{eq:least_square1} follows from Eq.~\eqref{eq:prior} and Eq.~\eqref{eq:likelihood}. The MAP estimation problem can therefore be reduced to a nonlinear least squares problem and can be solved with tools like Gauss-Newton or Levenberg-Marquardt.

When solving estimation and planning simultaneously, we encounter new measurements and/or cost information during online execution, thus changing the likelihood. A na\"{\i}ve approach to contending with this new information would be to resolve the problem in Eq.~\eqref{eq:least_square1} repeatedly. However, this is very inefficient and computationally expensive for an online setting. In the following sections, we formulate the inference problem on graphical models that allow us to exploit the underlying sparsity of the problem (Section~\ref{sec:fac}-\ref{sec:steap}), and then we use incremental inference techniques that allow us to iteratively update the solution only where needed resulting in a computationally efficient approach (Section ~\ref{sec:incinf}).


\section{Mapping, Estimation, and Planning with Factor Graphs}\label{sec:fac}

The MAP trajectory computation in Section~\ref{sec:MAP} can be executed efficiently by exploiting known structure in the problem. This is accomplished by representing the posterior distribution as a \emph{factor graph}. With a factor graph \citep{kschischang2001factor} any distribution can be \emph{factored} into a product of functions that is organized as a bipartite graph $G = \{{\Theta}, {F}, {E} \}$. This graph consists of variable nodes ${\Theta} \doteq \{\bm{\theta}_0, \dots , \bm{\theta}_N\}$, factor nodes ${F} \doteq \{ f_0, \dots , f_M \}$, and edges ${E}$ that connect the two types of nodes.

In our case, the variables are a set of instantaneous robot states along the trajectory, and the factors are conditional probability distributions on variable subsets ${\Theta}_i$ of ${\Theta}$. Therefore, we can write the posterior distribution as a product of the factors
\begin{equation}
p(\bm{\theta} | \mathbf{e}) \propto \prod \limits_{i=0}^M f_{i} ({\Theta}_i). \label{eq:factor_graph}
\end{equation}
The precision matrix of this distribution also encodes the connectivity in the graph. Consequently, a sparse factor graph structure yields a sparse precision matrix, which can be exploited to make the computation in Eq.~\eqref{eq:least_square1} efficient \citep{dellaert2006square}.

We can further write the posterior distribution as a product of prior factors and likelihood factors,
\begin{equation}
	p(\bm{\theta} | \mathbf{e}) \propto p(\bm{\theta})  p( \mathbf{e} | \bm{\theta}) \propto f^{prior}({\Theta}) f^{like}({\Theta}). \label{eq:fg}
\end{equation}
In the remainder of this section, we will use this general formulation to illustrate relationships between prior work in mapping, estimation, and planning. Then, we will extend this idea and connect it with our proposed work in the next section.

\paragraph{\bf SAM:} We begin with the smoothing and mapping (SAM) \citep{dellaert2006square} problem, an early work that uses factor graphs to address the state estimation and mapping problem in robotics. The goal is to estimate the full posterior trajectory in the past given all measurements. The factor graph used in SAM is 
\begin{equation}
p(\bm{\theta}_{est} | \mathbf{e}) \propto f^{prior} f^{meas},
\end{equation}
where $\bm{\theta}_{est}$ signifies the history portion of the trajectory to be estimated, $f^{prior} = f^{prior}(\bm{\theta}_0)$ is the prior on the first state, and $f^{meas}$ is the likelihood of all sensor measurements, which itself factors as
\begin{equation}
	f^{meas} = \prod_i f^{meas}_i({\Theta_i}).
\end{equation}
Unary measurement factors can refer to odometer, GPS or IMU measurements, while higher order factors on a subset of states (${\Theta_i}$) often represent landmark observations.

\paragraph{\bf STEAM:} Like SAM, simultaneous trajectory estimation and mapping (STEAM) \citep{barfoot2014batch,anderson2015full} addresses trajectory estimation problems. The key difference is that in STEAM, the trajectory is no longer treated as a discrete sequence of states ${\Theta}$, but rather a continuous-time trajectory sampled from a GP. The prior is a joint distribution on the full trajectory $f^{prior} = f^{gp}({\Theta})$, yielding a factor graph
\begin{equation}
p(\bm{\theta}_{est} | \mathbf{e}) \propto f^{gp} f^{meas}.
\end{equation}
\paragraph{\bf GPMP2:} GPMP2 \citep{Dong-RSS-16} is a probabilistic inference framework for solving planning problems. It utilizes the GP trajectory representation from STEAM to find collision free future trajectories that satisfy the GP prior. Unlike SAM and STEAM, however, the likelihood is not based on sensor measurements, but rather the likelihood of a trajectory being free from collision with obstacles. The collision factor is defined as
\begin{equation}
	f^{obs} = \prod_i f^{obs}_i(\bm{\theta}_i).
\end{equation}
A fixed start and goal state (this can also be an end effector goal in workspace) is also required in planning problem, and therefore can be incorporated in to the likelihood. So factors to fix start and goal configurations are also employed
\begin{equation}
	f^{fix} = f^{start}(\bm{\theta}_0) f^{goal}(\bm{\theta}_N).
\end{equation}
The full factor graph of GPMP2 is, therefore
\begin{equation}
p(\bm{\theta}_{plan} | \mathbf{e}) \propto f^{gp} f^{obs} f^{fix}
\end{equation}
where $\bm{\theta}_{plan}$ signifies the future portion of the trajectory to be planned.
\paragraph{\bf SLAP:} In real robotics applications, it is frequently the case that both estimation and planning problems must be solved. One  approach to tackling this problem is SLAP \citep{penny2014simultaneous,agha2015simultaneous}. Although previous work in this area does not employ factor graphs, we reformulate SLAP using them here to illustrate its relation to other problems. SLAP can be viewed as splitting the inference problem into two factor graphs, an estimation graph and a planning graph, defined by
\begin{align}
p(\bm{\theta}_{est} | \mathbf{e}) & \propto f^{prior} f^{meas}, \\
p(\bm{\theta}_{plan} | \mathbf{e}) & \propto f^{prior} f^{curr} f^{obs} f^{goal}.
\end{align}
If a continuous-time trajectory representation like GPs are employed, we can replace $f^{prior}$ with $f^{gp}$. SLAP solves the estimation graph first to find an estimate of the current state $f^{curr} (\bm{\theta}_{curr})$ and then uses it to initialize and solve the planning problem.

\paragraph{}
We summarize the factorization of these various problems in Table.~\ref{table:factorization}, with their factor graphs shown in Fig.~\ref{fig:fgs}.

\begin{table}[t]
	\begin{center}
		\caption{Summary of related problems.}
		\label{table:factorization}
		\centering
		\scalebox{0.93}{
			\begin{tabular}{|c|c|c|}
				\hline
				\bf{Method} & \bf{Problem solved} & \bf{Factorization}\\
				\hline
				SAM & Estimation + Mapping & $f^{prior} f^{meas}$ \\
				STEAM & Estimation + Mapping & $f^{gp} f^{meas}$\\
				GPMP2 & Planning & $f^{gp} f^{obs} f^{fix}$\\
				SLAP & Estimation + Planning & Estimation : $f^{prior} f^{meas}$\\
				&& Planning : $f^{prior} f^{obs} f^{fix}$\\
				STEAP & Estimation + Planning & $f^{gp} f^{meas} f^{obs} f^{fix}$\\
				\hline
		\end{tabular}}
	\end{center}
\end{table}


\section{Simultaneous Trajectory Estimation and Planning }\label{sec:steap}

In this paper, we present simultaneous trajectory estimation and planning (STEAP), where the task is to perform inference on the entire factor graph from start to goal at once, in contrast to SLAP, which would solve the estimation and planning graph sequentially. We optimize the full trajectory $\bm{\theta} = \bm{\theta}_{est} \cup \bm{\theta}_{plan}$ represented by the GP prior in Section~\ref{sec:traj_prior} given all sensor data and cost information collected in to a single likelihood. Compared to prior work discussed in the previous section, the likelihood here is interpreted more broadly to represent events than happen in the past and in the future, all together. The STEAP factor graph is defined as,
\begin{equation}
p(\bm{\theta} | \mathbf{e}) \propto f^{gp} f^{meas} f^{obs} f^{fix}.
\end{equation}
We define these factors in Section~\ref{sec:steap_factors}. Given the current time-step, the solutions to the estimation and planning problems automatically fall out. During online execution as new measurement or cost information is encountered, the likelihood, and by extension the factor graph, can be updated appropriately. Performing inference on the new graph then provides the updated estimation and replanning solutions. We explain this procedure with a simple toy example in Section~\ref{sec:example}. STEAP has the following major advantages:

(i) Optimization of a single graph allows information to flow between the two sub-graphs of estimation and planning, which is not possible with SLAP. This increases performance in both estimation and planning, and provides mutual benefit. The collision-free likelihood of both the past and the future part of the graph encourages the estimated past trajectory to remain in areas without obstacles, since a successfully traversed trajectory would not have passed through obstacles. This helps contend with noisy (or drops in) raw measurements and reduces trajectory estimation errors. Similarly, the trajectory estimation information corrects the estimate of the current robot position, providing feedback for the planned future trajectory. 

(ii) The number of variables in a STEAP factor graph does not change much during execution i.e. only a few factors are added in each step. This allows for very efficient incremental inference using the Bayes tree data structure \citep{kaess2011bayes}. Updating the solution with Bayes trees only requires a small fraction of the runtime compared to reoptimizing the full graph from scratch. Additional discussion of incremental inference using the Bayes tree is in Section~\ref{sec:incinf}.

\begin{figure*}[!ht]
	\centering
	\begin{minipage}[t]{0.40\columnwidth}
		\vspace{0pt}
		\centering
		\includegraphics[width=1\linewidth]{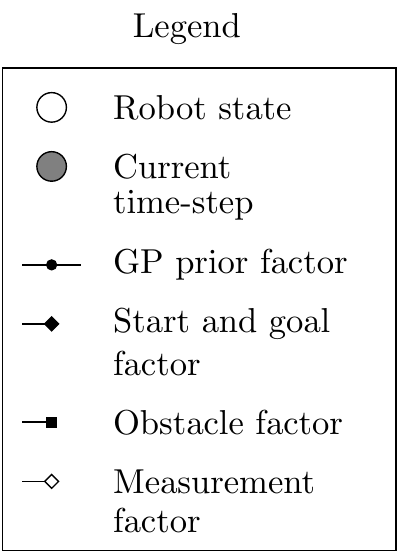}
	\end{minipage}
	\hfill
	\begin{minipage}[t]{1.0\columnwidth}
		\vspace{0pt}
		\centering
		\includegraphics[width=1\linewidth]{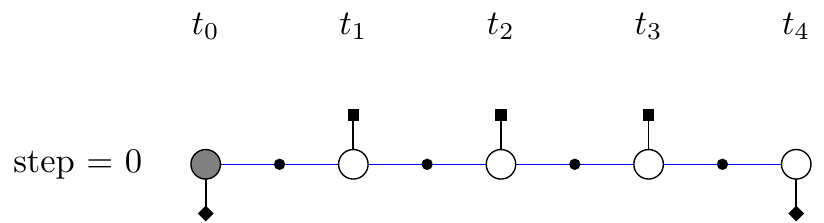}
		\par\bigskip
		\includegraphics[width=1\linewidth]{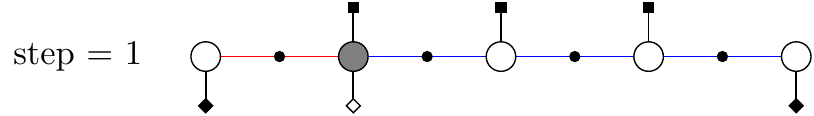}
		\par\bigskip
		\includegraphics[width=1\linewidth]{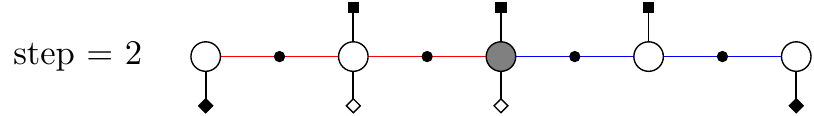}
		\par\bigskip
		\includegraphics[width=1\linewidth]{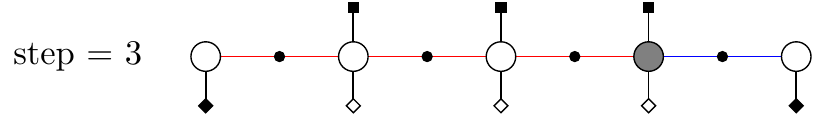}
		\par\bigskip
		\includegraphics[width=1\linewidth]{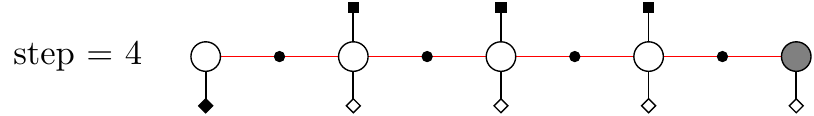}
	\end{minipage}
	\hfill
	\begin{minipage}[t]{0.6\columnwidth}
		\vspace{24pt}
		\includegraphics[trim={2cm 2cm 1.5cm 2cm},clip,width=1\linewidth]{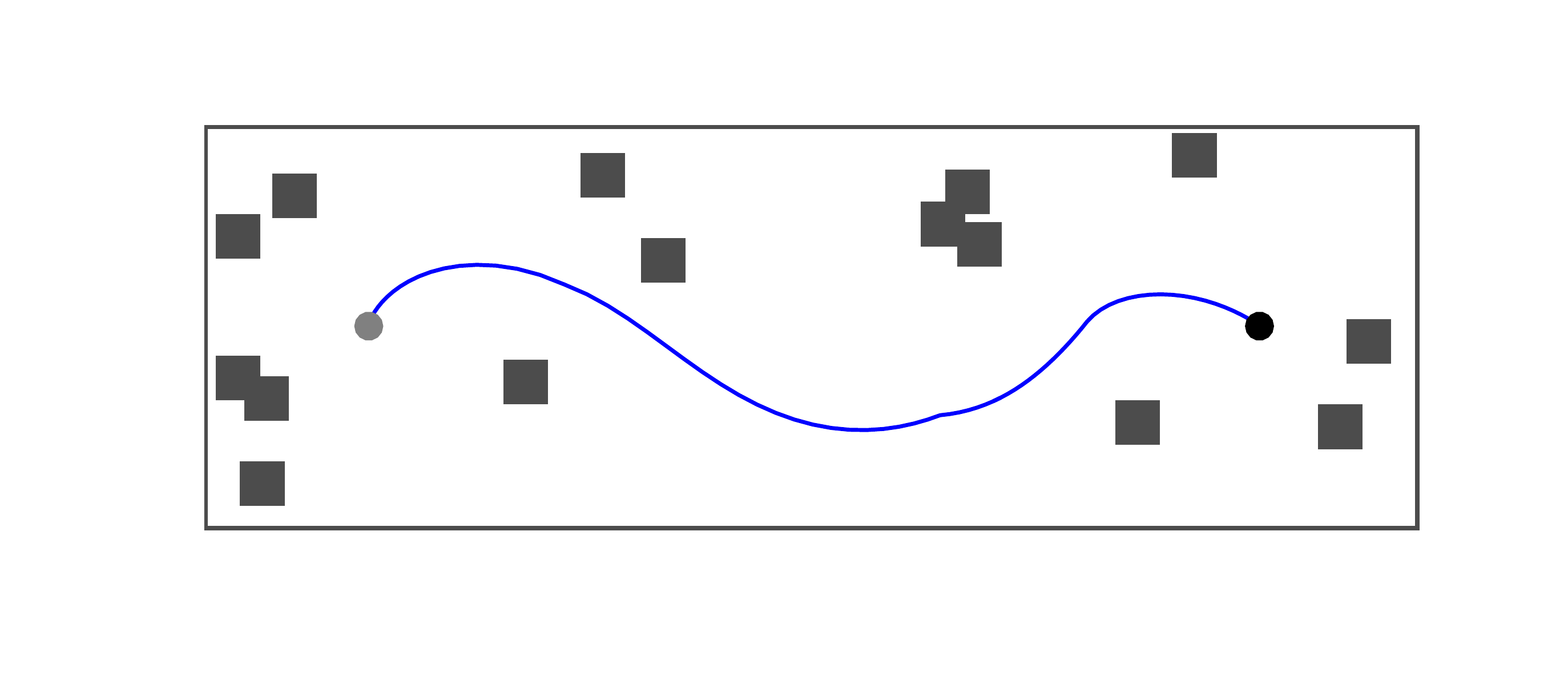}
		\includegraphics[trim={2cm 2cm 1.5cm 2cm},clip,width=1\linewidth]{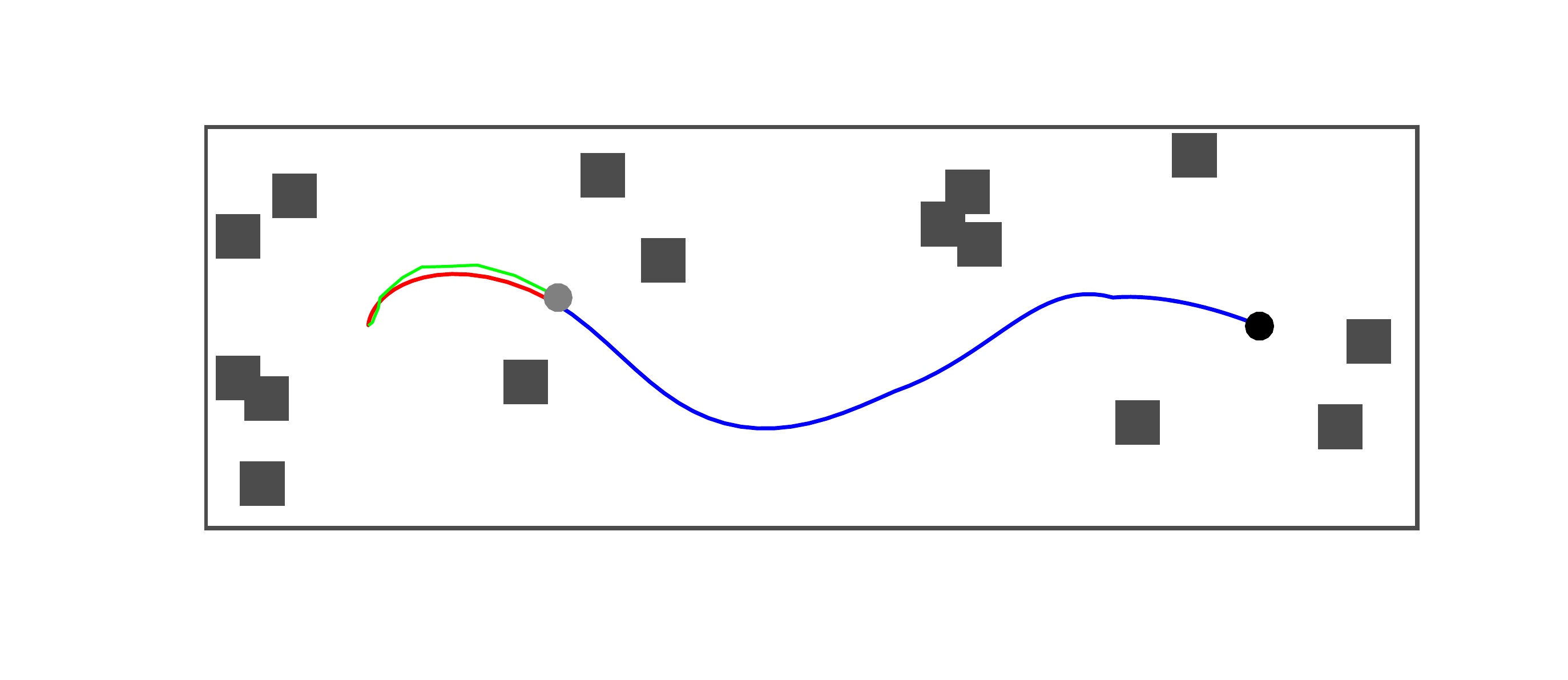}
		\includegraphics[trim={2cm 2cm 1.5cm 2cm},clip,width=1\linewidth]{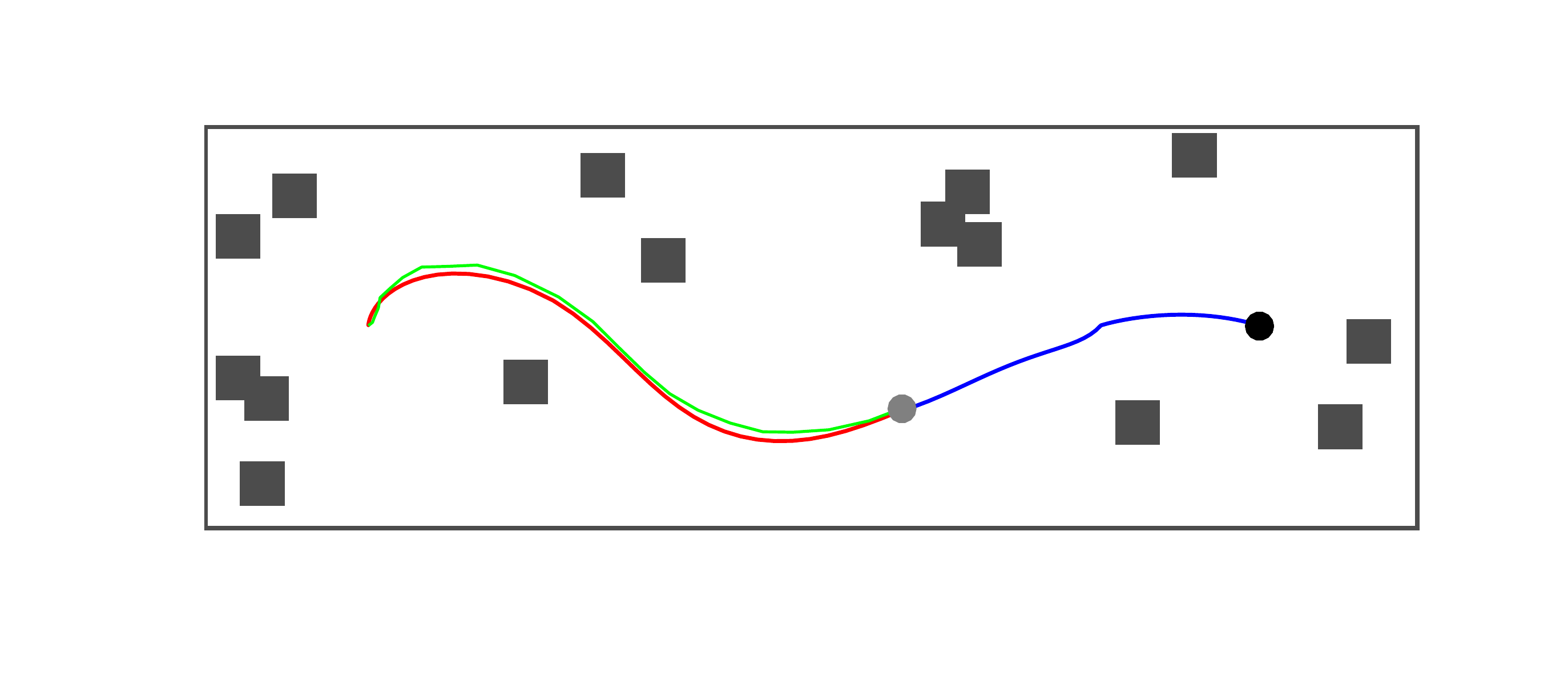}
		\includegraphics[trim={2cm 2cm 1.5cm 2cm},clip,width=1\linewidth]{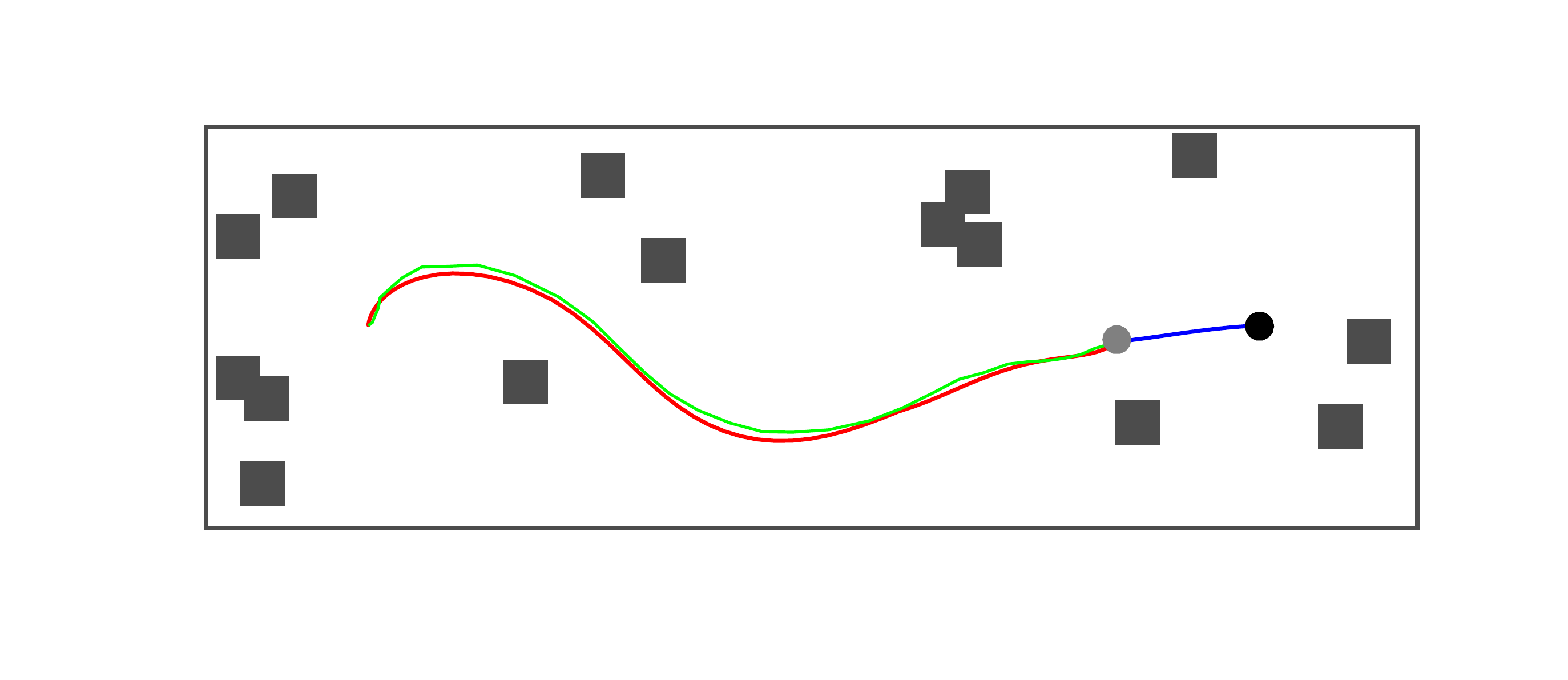}
		\includegraphics[trim={2cm 2cm 1.5cm 2cm},clip,width=0.99\linewidth]{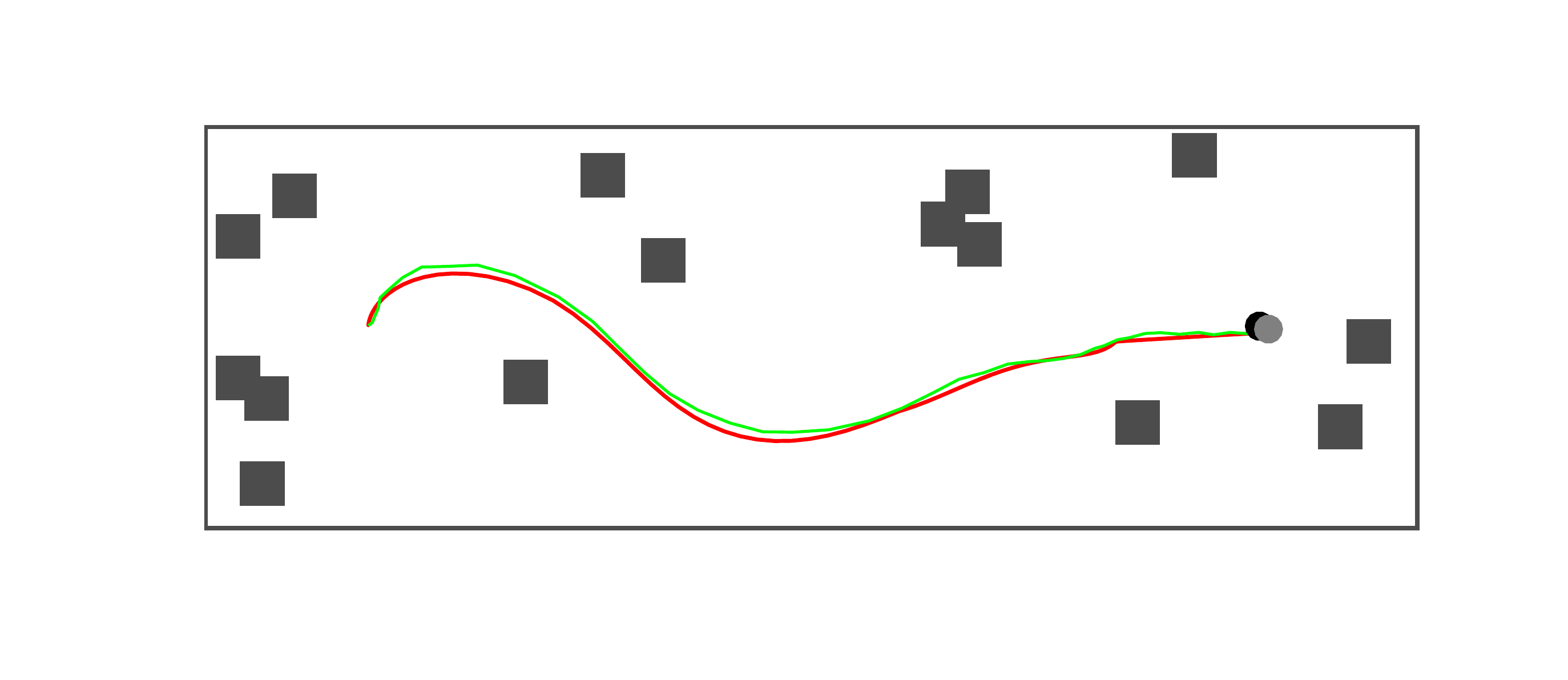}
	\end{minipage}
	\protect\caption{A simple example illustrates STEAP using a robot (gray) that navigates to the goal (black circle) while avoiding obstacles. At each step the right side shows the environment with ground-truth (green), estimated (red), and replanned (blue) trajectories. The left side shows the corresponding factor graph. See text for details.
		\label{fig:example}}
\end{figure*}

\subsection{STEAP factor definitions}\label{sec:steap_factors}

\subsubsection{The Gaussian process prior factor}\label{sec:prior}

A Gaussian RBF kernel defines a prior distribution of trajectories with no pairwise independences. In other words, all states are connected to a single GP prior factor, $f^{gp} = f^{gp}({\Theta})$. This prior cannot be factored, and destroys the problem's sparsity, making inference computationally expensive.
However, in the context of STEAM problems, \citet{barfoot2014batch} showed that certain types of GP priors generated by linear time varying (LTV) stochastic differential equations (SDEs), are sufficient to model Markovian robot trajectories. These priors are highly structured, and factor according to
\begin{equation}
f^{gp} = \prod_i f^{gp}_i(\bm{\theta}_i, \bm{\theta}_{i+1})
\end{equation}
where any GP prior factor connects to only its two neighboring states, forming a (Gauss-Markov) chain. This is shown in Fig.~\ref{fig:fgs} (a) where states (white circle) form a chain by connecting to GP prior factors (black circle). In GPMP2 \citep{Dong-RSS-16}, the GP prior on trajectories is generated by a LTV-SDE defined on a vector space (Fig.~\ref{fig:fgs} (b)). GP priors have also been formulated with non-linear SDEs \citep{anderson2015batch} and on the SE(3) Lie group \citep{anderson2015full}. 

If the robot configuration is in vector space $\mathbb{R}^n$, similar to GPMP2, STEAP can use the GP prior defined in \citep{barfoot2014batch}. But we develop STEAP for mobile manipulators that have their configuration space defined by a Lie group product $\bm{\theta}_{i} \in$ SE(2)$\times \mathbb{R}^n$ where $n$ is the degree-of-freedom of the arm and the SE(2) Lie group defines a planar translation and rotation (yaw) for the mobile base. We employ a constant velocity i.e. noise-on-acceleration model to define a non-linear SDE that generates our GP prior. See Section~\ref{sec:gp_prior} for details about the GP prior.

\subsubsection{Obstacle factor}
All obstacle factors are constructed similar to GPMP2 \citep{Dong-RSS-16} except that they are defined for the Lie group configuration space. The obstacle factors evaluate collision cost using a hinge loss function and a signed distance field of the environment. See \citep{Dong-RSS-16} for details.

\subsubsection{Start and goal factor}
These are multivariate Gaussian factors
\begin{align}
	f^{start}(\bm{\theta}_0) = \exp\bigg\{-\frac{1}{2}\parallel \bm{\theta}_0 - \bm{\theta}_{start} \parallel^2_{\Sigma_{fix}}\bigg\} \\
	f^{goal}(\bm{\theta}_N) = \exp\bigg\{-\frac{1}{2}\parallel \bm{\theta}_N - \bm{\theta}_{goal} \parallel^2_{\Sigma_{fix}}\bigg\}
\end{align}
with the mean as the start or goal and a small covariance $\Sigma_{fix}$, and are used to tie down the trajectory at the start and goal locations. When the trajectory has finished execution, the goal factor is replaced with the pose measurement factor so that the final posterior update gives the final trajectory estimate.

\subsubsection{Measurement factor}
There are many types of sensors that provide different measurements, and thus many types of measurement factors have been proposed by the SLAM community \citep{dellaert2006square}. For example, measurements from an inertial measurement unit (IMU) can be incorporated into the factor graph with pre-integrated IMU factors~\citep{Forster-RSS-15}, and visual landmark measurements from a camera can be incorporated with Schur complement factors~\citep{carlone2014eliminating}.

For the sake of simplicity, in the remainder of this paper we use a multivariate Gaussian measurement factor for the state measurement 
\begin{equation}
f^{meas}_i(\bm{\theta}_i) = \exp\bigg\{-\frac{1}{2}\parallel \bm{\theta}_i - \bm{\mu}^{meas}_i \parallel^2_{\Sigma_{meas}}\bigg\}
\end{equation}
where the measurement queried from sensors has mean $\bm{\mu}^{meas}_i$ with covariance $\Sigma_{meas}$. The multivariate Gaussian is a typical noise assumption for many sensors. For example, GPS can provide coordinate and velocity measurements in Cartesian space with covariance~\citep{leandro2005empirical}, and 2D/3D laser scanners can provide a raw localization with covariance from a 2D/3D point-cloud with the ICP algorithm~\citep{censi2007accurate}. Note that, although we only use multivariate Gaussian measurement factors in our examples and evaluations, the STEAP framework is general and can incorporate any type of measurement factors.

\subsection{A STEAP example}\label{sec:example}
We use an example, illustrated in Fig.~\ref{fig:example}, to describe how \algo works using Algorithm~\ref{alg:steap} that is complemented by the block diagram in Fig.~\ref{fig:block}. Note that the small size of the graph in Fig.~\ref{fig:example} is just for illustration, in practice our approach can handle much larger graphs (see Section \ref{sec:eval} for the graph sizes used in our experiments). In this example, a robot with stochastic dynamics starts at time-step $t_0$ and needs to reach the goal at time-step $t_4$ while avoiding any obstacles.

First, we construct a factor graph that will reflect the prior distribution. A small, sparse set of robot states are connected via GP factors that collectively form the prior distribution of a continuous-time trajectory. Then, we add a start and a goal factor with a small covariance (to tie the trajectory down at the start and goal) and obstacle factors. In practice, there are also multiple binary obstacle factors present between any two states (omitted here for clarity) that use GP interpolation to project the cost between any two states back on to those states and allow the trajectory to stay sparse but still reason about obstacles between the sparse states (see \citep{Dong-RSS-16} for details). The start, goal and obstacle factors together form the likelihood. We can find the mode of the posterior shown in blue at the top level of Fig.~\ref{fig:example}, which is inherently a special case of our approach providing the solution to the GPMP2 motion planning problem, since no measurement factors are present and there is no state estimation yet at this step.

\begin{algorithm}[!t]
	\caption{STEAP}\label{alg:steap}
	\hspace*{\algorithmicindent} $\bm{\theta}$: trajectory, $f$: factors, ${T}$: Bayes tree
	\begin{algorithmic}[1]
		\STATE $\bm{\theta}_{init}$ = initializeTrajectory()
		\STATE $FG$ = createFactorGraph($f^{gp}$, $f^{obs}$, $f^{fix}$)
		\STATE $\bm{\theta}$ = inference($FG$, $\bm{\theta}_{init}$)
		\STATE ${T}$ = createBayesTree($FG$, $\bm{\theta}$)
		\FOR {$i=0$ \TO $N-1$}
		\STATE $\theta_{i:i+1}$ = interpolateGP($\bm{\theta}$, $i$, $i+1$, resolution)
		\IF {collisionFree($\theta_{i:i+1}$)}
		\STATE execute($\theta_{i:i+1}$)
		\STATE $f^{meas}_{i+1}$ = localize()
		\STATE $\bm{\theta}$, ${T}$ = incrementalInference($\bm{\theta}$, $f^{meas}_{i+1}$, ${T}$)
		\ELSE
		\RETURN failure
		\ENDIF
		\ENDFOR
		\RETURN success
	\end{algorithmic}
\end{algorithm}	

Next, the planned solution between $t_0$ and $t_1$ is upsampled to a desired resolution with GP interpolation, checked for safety, and then executed on the robot. The ground-truth trajectory is illustrated in green. Since the system is stochastic, execution is noisy. We make an observation to generate a measurement factor and insert it into the graph at $t_1$. This new factor is combined with the old likelihood to produce the updated likelihood. Using the Bayes tree to efficiently organize computation, we generate a new MAP solution. Note that, in this case, the factor graph is changed by adding only one measurement factor, so the incremental inference performed using the Bayes tree will be very fast. The red portion of the trajectory is an estimate of the trajectory traversed by the robot until current time-step $t_1$ and the blue portion of the trajectory is the replanned solution to the goal. This whole process is then repeated (steps are shown from top to bottom in Fig.~\ref{fig:example}) until the robot reaches the goal at $t_4$. At $t_4$ again we have a special case of our approach that provides a solution to the trajectory estimation problem (STEAM), but with extra obstacle factors.


\section{Incremental Inference with the Bayes Tree Data Structure}\label{sec:incinf}

In Section~\ref{sec:MAP}, we discussed how to solve the MAP inference problem as non-linear least squares optimization. But one significant drawback of using non-linear optimization like Gauss-Newton or Levenberg-Marquardt to solve inference on factor graphs is that they are iterative methods, and with every iteration the problem must be completely linearized and resolved (the cost on every factor will be evaluated and every variable is updated), even if the factor graph is mostly unchanged. 

To reduce these redundant calculations, \citet{kaess2011isam2} proposed efficient incremental updates of non-linear least square problems with the \emph{Bayes tree} data structure. When re-solving a graph with only minor changes (in variables or factors), only the parts of the Bayes tree associated with the changes will be updated, leaving most of the Bayes tree unchanged. By updating the solution in this \emph{incremental} manner, the efficiency of inference is significantly improved.

Since only a very small portion of the STEAP factor graph changes (few variables with new measurement factors) at each time-step, we convert the factor graph into a Bayes tree, and update the tree incrementally. By utilizing this efficient incremental inference technique we get a significant performance boost, and easily achieve real-time performance, as illustrated in our experiments.

In this section, we first give a brief overview of the Bayes tree and its relation to factor graphs, discuss how to perform incremental inference on Bayes tree, and then give a detailed example to show how to use a Bayes tree to perform incremental inference for a STEAP problem. Readers are encouraged to refer to the iSAM2 paper \citep{kaess2011isam2} for a more detailed and general explanation.

\subsection{Building a Bayes tree from a factor graph}\label{sec:bayestree}

\begin{figure*}[t]
	\begin{centering}
		\begin{subfigure}[b]{0.32\textwidth}
			\centering
			\includegraphics[width=1\linewidth]{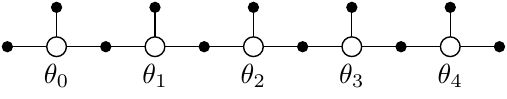}
			\caption{Factor graph}
		\end{subfigure}
		\hfill
		\begin{subfigure}[b]{0.32\textwidth}
			\centering
			\includegraphics[width=1\linewidth]{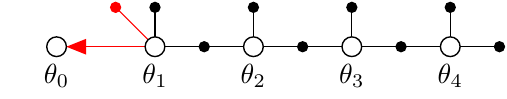}
			\caption{Eliminating $\theta_0$}
		\end{subfigure}	
		\hfill
		\begin{subfigure}[b]{0.32\textwidth}
			\centering
			\includegraphics[width=1\linewidth]{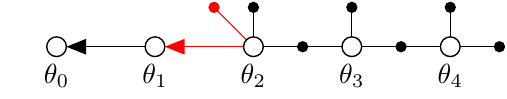}
			\caption{Eliminating $\theta_1$}
		\end{subfigure}
	\end{centering}
	\begin{centering}
		\begin{subfigure}[b]{0.32\textwidth}
			\centering
			\includegraphics[width=1\linewidth]{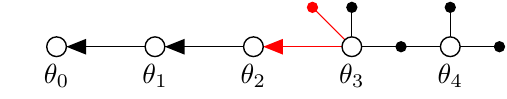}
			\caption{Eliminating $\theta_2$}
		\end{subfigure}
		\hfill
		\begin{subfigure}[b]{0.32\textwidth}
			\centering
			\includegraphics[width=1\linewidth]{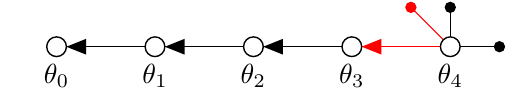}
			\caption{Eliminating $\theta_3$}
		\end{subfigure}	
		\hfill
		\begin{subfigure}[b]{0.32\textwidth}
			\centering
			\includegraphics[width=1\linewidth]{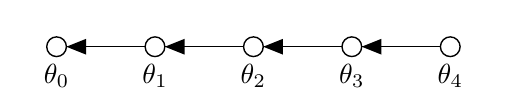}
			\caption{Eliminating $\theta_4$ and final Bayes net}
		\end{subfigure}
	\end{centering}
	\caption{Example of applying variable elimination on a planning factor graph. Red arrows/factors indicate the parts that change in Bayes net/factor graph respectively at step 5 of Algorithm~\ref{alg:elimination}.}
	\label{fig:elimination_example}
\end{figure*}

A Bayes tree is a directed tree-structured graphical model which is derived from a Bayes net that has very close relation to junction tree. Both the Bayes tree's and junction tree's nodes are \emph{cliques} of a Bayes net, but the Bayes tree is \emph{directed} to reflect the conditional relations in factored probability density.

Before we officially define the Bayes tree, we first introduce the \emph{variable elimination} algorithm~\citep{dellaert2006square}, which converts a factor graph into a \emph{Bayes net}. For a given factor graph, we first choose an ordering of variables. Although any variable ordering works for the explanation here, different orderings generate Bayes nets with different numbers of edges, and, in general, a smaller number of edges is better for reducing computation. Choosing the optimal ordering is a NP-hard problem. To contend with this problem, several approximation heuristics have been proposed. We use COLAMD~\citep{davis2004column} to estimate a close-to-optimal ordering.

Given a factor graph and a variable ordering, we eliminate each variable by Algorithm~\ref{alg:elimination} and factorize the probability density over all variables to
\begin{equation}
p(\theta) = \prod_j p(\theta_j | S_j),\label{eq:bayes_net}
\end{equation}
where $S_j \subset \theta$ is the separator of $\theta_j$. Note that the factorized probability density in Eq.~\eqref{eq:bayes_net} meets the conditional dependencies of a Bayes net, so, by elimination, we convert a factor graph to a Bayes net. Fig.~\ref{fig:elimination_example} shows an example of running elimination Algorithm~\ref{alg:elimination} on a GPMP2 planning factor graph with reverse variable ordering, which is actually the optimal variable ordering in this case.

\begin{algorithm}[!t]
\caption{Variable elimination of factor graph}\label{alg:elimination}
\begin{algorithmic} [1]
\FORALL {$\theta_j$, in ordering,}
    \STATE Remove all factors $f_i$ connected to $\theta_j$ from factor graph, define $S_j = \{$all variables involved in all $f_i\}$.
    \STATE $f_j(\theta_j, S_j) = \prod_i f_i(\theta)$.
    \STATE Factorize $f_j(\theta_j, S_j) = p(\theta_j|S_j)f_{new}(S_j)$.
    \STATE Add $p(\theta_j|S_j)$ in Bayes net, add $f_{new}(S_j)$ back in factor graph.
\ENDFOR
\end{algorithmic}
\end{algorithm}

\begin{figure}[t]
	\begin{centering}
		{\includegraphics[width=0.9\columnwidth]{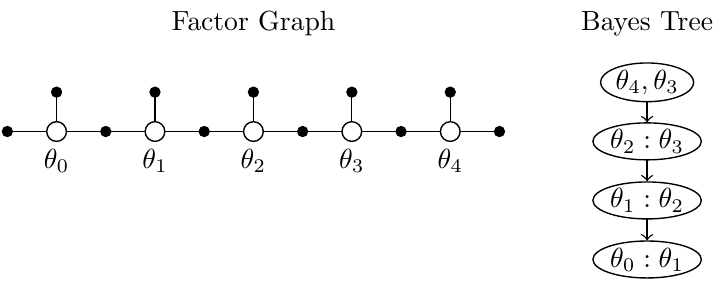}}
		\par\end{centering}
	\protect\caption{Example of a Bayes Tree of a planning factor graph. 
		\label{fig:bayes_tree_example}}
\end{figure}

For a Bayes net generated by Algorithm~\ref{alg:elimination}, we extract all cliques $C_k$ from the Bayes net and build a Bayes tree by defining each node by one clique $C_k$. For each node of the Bayes tree, we further define the conditional density $p(F_k|S_k)$, where $S_k$ is the \emph{separator variables} $S_k$, by $S_k = C_k \cap \Pi_k$ intersecting between $C_k$ and $C_k$'s parent node $\Pi_k$, and the \emph{frontal variables} $F_k$, by $F_k = C_k \backslash S_k$. The clique is written as $C_k = F_k : S_k$. The probability density of a Bayes tree is defined by the joint density of all nodes
\begin{equation}
p(\theta) = \prod_k p_{C_k}(F_k | S_k).\label{eq:bayes_tree}
\end{equation}
The algorithm to convert a Bayes net to a Bayes tree is summarized in Algorithm~\ref{alg:bayes_tree}, and the Bayes tree example of the toy planning factor graph from Fig.~\ref{fig:elimination_example} is illustrated in Fig.~\ref{fig:bayes_tree_example}. Here, the example is straightforward: the Bayes net has 4 cliques, $\{\theta_0, \theta_1\}$, $\{\theta_1, \theta_2\}$, $\{\theta_2, \theta_3\}$ and $\{\theta_3, \theta_4\}$, so the Bayes tree has 4 nodes.

\begin{algorithm}[!t]
\caption{Creating a Bayes tree from a Bayes net}\label{alg:bayes_tree}
\begin{algorithmic} [1]
\FORALL {$p(\theta_j|S_j)$, in \emph{reverse} ordering,}
    \IF {$S_j=\emptyset$}
        \STATE Start a tree with a root clique $C_r$ with $p_{C_r} = p(\theta_j)$, $F_r = \{\theta_j\}$.
    \ELSE
        \STATE Find the parent clique $C_p$ that contains the first eliminated variable in $S_j$ is in $F_p$.
        \IF {$F_p \cap S_p \subseteq S_j$}
            \STATE Add $p(\theta_j|S_j)$ in $p_{C_p}$, add $\theta_j$ in $F_p$, and add $S_j$ in $S_p$.
        \ELSE
            \STATE Add a new clique $C'$ in tree with $p_{C'} = p(\theta_j|S_j)$, $F' = \{\theta_j\}, S' = S_j$ as a child node of $C_p$.
        \ENDIF
    \ENDIF
\ENDFOR
\end{algorithmic}
\end{algorithm}

\subsection{Incremental inference on Bayes tree}\label{sec:bayestree_incinf}

Given a Bayes tree generated from a factor graph, we can efficiently add new factors and perform incremental inference. One of the most important properties of the Bayes tree is that if a factor, involving a variable $\theta_i$, is added in the Bayes tree, only the cliques between the cliques containing $\theta_i$ and the root of the tree (assume the root of the tree is at the top) will be affected. The sub-tree below the cliques containing $\theta_i$ will remain unchanged during incremental inference. This means that by updating the Bayes tree a large part of the computation that is redundant and involves only untouched cliques, can be prevented compared to solving a full new non-linear least squares optimization problem.

\begin{figure}[!t]
\centering
\begin{subfigure}[b]{0.6\columnwidth}
\centering
\includegraphics[width=1\linewidth]{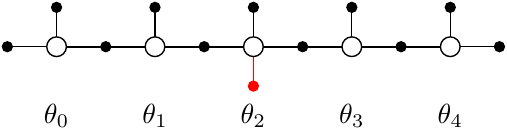}
\caption{Add a factor to factor graph}
\end{subfigure}
\begin{subfigure}[b]{0.9\columnwidth}
\centering
\includegraphics[width=1\linewidth]{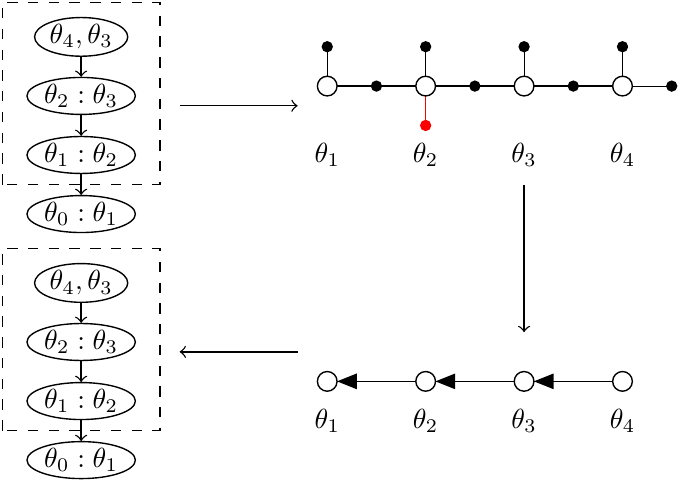}
\caption{Adding a factor to a Bayes tree}
\end{subfigure}
\caption{An example of how to add factors and perform incremental inference on a Bayes tree. (a) The inference problem illustrated as a factor graph. The added factor is displayed in red. (b) Steps to add the red factor into Bayes tree.}\label{fig:update_bayes_tree_example}
\end{figure}

The procedure to update a Bayes tree with new factors is stated in Algorithm~\ref{alg:incr_inf}. Given a set of new factors, we first find all the cliques which contain the variables involved in those new factors, and reinterpret the sub-tree as a factor graph. After adding new factors in the factor graph, we perform elimination on the factor graph to get the corresponding Bayes net, and further convert the Bayes net in to a Bayes tree. Finally, we attach the untouched sub-tree to the updated sub-tree, to get the final updated tree. The procedure is further explained by a toy example in Fig.~\ref{fig:update_bayes_tree_example}. In this example we add a unary factor to a planning factor graph at $\theta_2$. As explained in the previous sub-section, we have the Bayes net of the factor graph in Fig.~\ref{fig:update_bayes_tree_example}(b), where the dashed boxes mark the affected part of the tree. We see that the clique $\theta_0:\theta_1$ remains untouched during the whole update procedure.

\begin{algorithm}[!t]
\caption{Incremental inference on a Bayes tree}\label{alg:incr_inf}
\begin{algorithmic} [1]
\REQUIRE {Bayes tree ${T}$, add factors ${F}$}.
\STATE Remove top of ${T}$, reinterpret as factor graph.
\STATE Add ${F}$ to factor graph.
\STATE Eliminate factor graph to Bayes net, then to Bayes tree.
\STATE Attach unchanged sub-tree to updated sub-tree.
\RETURN {Updated Bayes tree ${T}^*$}.
\end{algorithmic}
\end{algorithm}

\subsection{Using the Bayes tree in STEAP}

\begin{figure}[t]
\centering
\includegraphics[width=0.9\columnwidth]{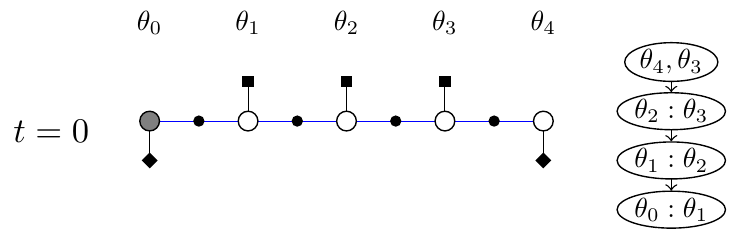}
\vspace{1mm}
\includegraphics[width=0.9\columnwidth]{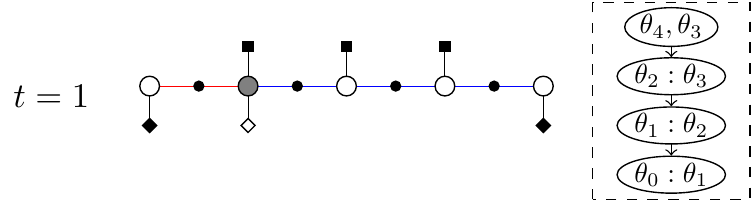}
\vspace{1mm}
\includegraphics[width=0.9\columnwidth]{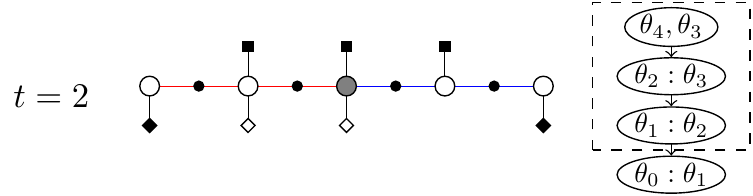}
\vspace{1mm}
\includegraphics[width=0.9\columnwidth]{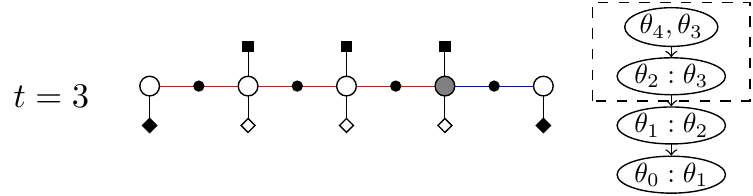}
\vspace{1mm}
\includegraphics[width=0.9\columnwidth]{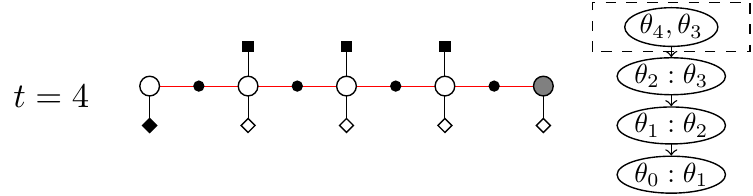}
\caption{The Bayes tree of the STEAP example (red and blue are the estimation and planning parts of the graph respectively) from Section~\ref{sec:example}. At $t=0$ the Bayes tree is built from the factor graph shown on the left, and from $t=1$ to $t=4$ the incremental inference is performed on the tree, with affected sub-trees marked by dashed boxes.}
\label{fig:steap_bayes_tree}
\end{figure}

We are now ready to discuss how the Bayes tree can be used to speed up STEAP. At the beginning of the STEAP algorithm, the factor graph is the same as the planning factor graph in GPMP2~\citep{Dong-RSS-16}. A Bayes tree is constructed from the factor graph using the approach discussed in Section~\ref{sec:bayestree}. After construction of the Bayes tree, the factor graph is no longer needed or maintained, since all remaining steps are performed directly on the Bayes tree. When the Bayes tree needs to be updated with new measurement factors, we update the tree and perform incremental inference using the approach described in Section~\ref{sec:bayestree_incinf}. An example is shown in Fig.~\ref{fig:steap_bayes_tree} that corresponds to the STEAP example in Section~\ref{sec:example}.

As mentioned in Section~\ref{sec:steap}, one of the advantages of STEAP, compared to other trajectory estimation and planning approaches, is that it can use Bayes trees to gain a significant speed up while performing incremental inference. In the example in Fig.~\ref{fig:steap_bayes_tree}, we see that for each step of STEAP, only part of the Bayes tree is updated leaving the remaining tree unchanged. Therefore, incremental inference is more efficient while maintaining similar accuracy compared to batch inference, which needs to perform all the steps (linearization, solving linear systems, etc.) on the full graph. We also gain additional efficiency improvement from the iSAM2~\citep{kaess2011isam2} implementation of the Bayes tree (which we use) that avoids excessive linearization operations by fluid relinearization. We will see in the experimental section how the Bayes tree and iSAM2 improve the efficiency of STEAP compared to other approaches.

\section{GP Priors}\label{sec:gp_prior}

\subsection{GP priors on vector space}

GP priors on vector-valued system states $\bm{\theta}(t)$ can be generated by linear time-varying stochastic differential equations (LTV-SDEs)~\citep{barfoot2014batch}
\begin{equation}
\dot{\bm{\theta}}(t) = \mathbf{A}(t)\bm{\theta}(t) + \mathbf{u}(t) + \mathbf{F}(t)\mathbf{w}(t), \label{eq:LTV-SDE}
\end{equation}
where $\mathbf{u}(t)$ is the known system control input, $\mathbf{w}(t)$ is white process noise, and both $\mathbf{A}(t)$ and $\mathbf{F}(t)$ are time-varying system matrices. The white process noise is represented by 
\begin{equation}
\mathbf{w}(t) \sim GP(\mathbf{0}, \mathbf{Q}_C\delta(t-t')), 
\end{equation}
where $\mathbf{Q}_C$ is the power-spectral density matrix, which is a hyperparameter~\citep{barfoot2014batch}, and $\delta(t-t')$ is the Dirac delta function. The mean and covariance of the GP is computed by taking the first and second order moments of the solution to Eq.~\eqref{eq:LTV-SDE}
\begin{align}
\bm{\mu}(t) &= \bm{\Phi}(t,t_0) \bm{\mu}_0 + \int_{t_0}^{t} \bm{\Phi}(t,s) \mathbf{u}(s) ds\\
\mathbf{{K}}(t,t') &=  \bm{\Phi}(t,t_0) \mathbf{{K}}_0 \bm{\Phi}(t',t_0)^\top+ \nonumber \\
&\int_{t_0}^{\min(t,t')} \bm{\Phi}(t,s) \mathbf{F}(s) \mathbf{Q}_C \mathbf{F}(s)^\top \bm{\Phi}(t',s)^\top ds
\end{align}
where $\bm{\mu}_0$ is the initial mean value of first state, $\mathbf{{K}}_0$ is the covariance of first state, and $\bm{\Phi}(t, s)$ is transition matrix.

\subsubsection{Constant velocity GP prior}

The \emph{constant-velocity} GP prior is generated by a LTV-SDE with a noise-on-acceleration model
\begin{equation}
\ddot{\mathbf{p}}(t) = \mathbf{w}(t), \label{eq:const-vel-LTV-SDE}
\end{equation} 
where $\mathbf{p}(t)$ is the $N$-dimensional vector-valued position (or pose) variable of trajectory, if the system has $N$ degrees of freedom. To convert this prior into the LTV-SDE form of Eq.~\eqref{eq:LTV-SDE}, a Markov system state variable is defined
\begin{equation}
\bm{\theta}(t) \doteq \begin{bmatrix} \mathbf{p}(t) \\ \dot{\mathbf{p}}(t) \end{bmatrix},
\end{equation}
The prior in Eq.~\eqref{eq:const-vel-LTV-SDE} then can easily be converted into a LTV-SDE in Eq.~\eqref{eq:LTV-SDE} by defining
\begin{equation}
\mathbf{A}(t) = \begin{bmatrix} \mathbf{0} & \mathbf{1} \\ 
\mathbf{0} & \mathbf{0} \end{bmatrix},\quad \mathbf{u}(t) = \mathbf{0},\quad 
\mathbf{F}(t) = \begin{bmatrix} \mathbf{0} \\ \mathbf{1} \end{bmatrix}. 
\label{eq:ltv-sde_constv}
\end{equation}

\subsubsection{GP prior factor}

Given the LTV-SDE formulation defined in Eq.~\eqref{eq:ltv-sde_constv}, we define the GP factor between any two states at $t_i$ and $t_{i+1}$ by~\citep{barfoot2014batch}
\begin{align}\label{eq:gp_prior_factor_vec}
f^{gp}_i(\bm{\theta}_i, \bm{\theta}_{i+1}) = \exp \bigg\{\hspace{-1mm} - \frac{1}{2} \| \bm{\Phi}(t_{i+1}, t_{i})\bm{\theta}_{i} - \bm{\theta}_{i+1} \|^{2}_{\mathbf{Q}_{i,i+1}} \hspace{-1mm} \bigg\}
\end{align}
where
\begin{equation}
\bm{\Phi}(t, s) = \begin{bmatrix} \mathbf{1} & (t-s) \mathbf{1} \\
\mathbf{0} & \mathbf{1} \end{bmatrix}, 
\mathbf{Q}_{i,i+1} = \begin{bmatrix} \frac{1}{3} \Delta t_i^3 \mathbf{Q}_C &
\frac{1}{2} \Delta t_i^2 \mathbf{Q}_C \\ 
\frac{1}{2} \Delta t_i^2 \mathbf{Q}_C &
\Delta t_i \mathbf{Q}_C \end{bmatrix}.
\end{equation}

\subsubsection{Constant time GP interpolation}

The posterior mean of the trajectory at \emph{any} time $\tau$ can be approximated by Laplace's method and expressed in terms of the current trajectory ${\bm{\theta}}$ at time points $\bm t$~\citep{barfoot2014batch}:
\begin{align}\label{eq:interp}
{\bm{\theta}}(\tau) = {\bm{\mu}}(\tau) + \mathbf{{K}}(\tau, {\bm t})\mathbf{{K}}^{-1}({\bm \theta}-\bm{\mu}).
\end{align}
Although the interpolation in Eq.~\eqref{eq:interp} na\"{\i}vely requires $O(N)$ operations, with structured kernels ${\bm{\theta}}(\tau)$ can be computed in as fast as $O(1)$~\citep{barfoot2014batch,Dong-RSS-16}.

\begin{figure*}[t]
	\centering
	\begin{minipage}[t]{0.18\textwidth}
		\caption{Block diagram of our framework showing all the components and how they interact. Blue boxes are modules and gray boxes are data. Sensor measurements flow from the \texttt{Robot Module} to the \texttt{Localization Module}. This block diagram is also complemented by Algorithm~\ref{alg:steap}.}
		\label{fig:block}
	\end{minipage}
	\hfill
	\begin{minipage}[t][][b]{0.78\textwidth}
		\includegraphics[width=\textwidth]{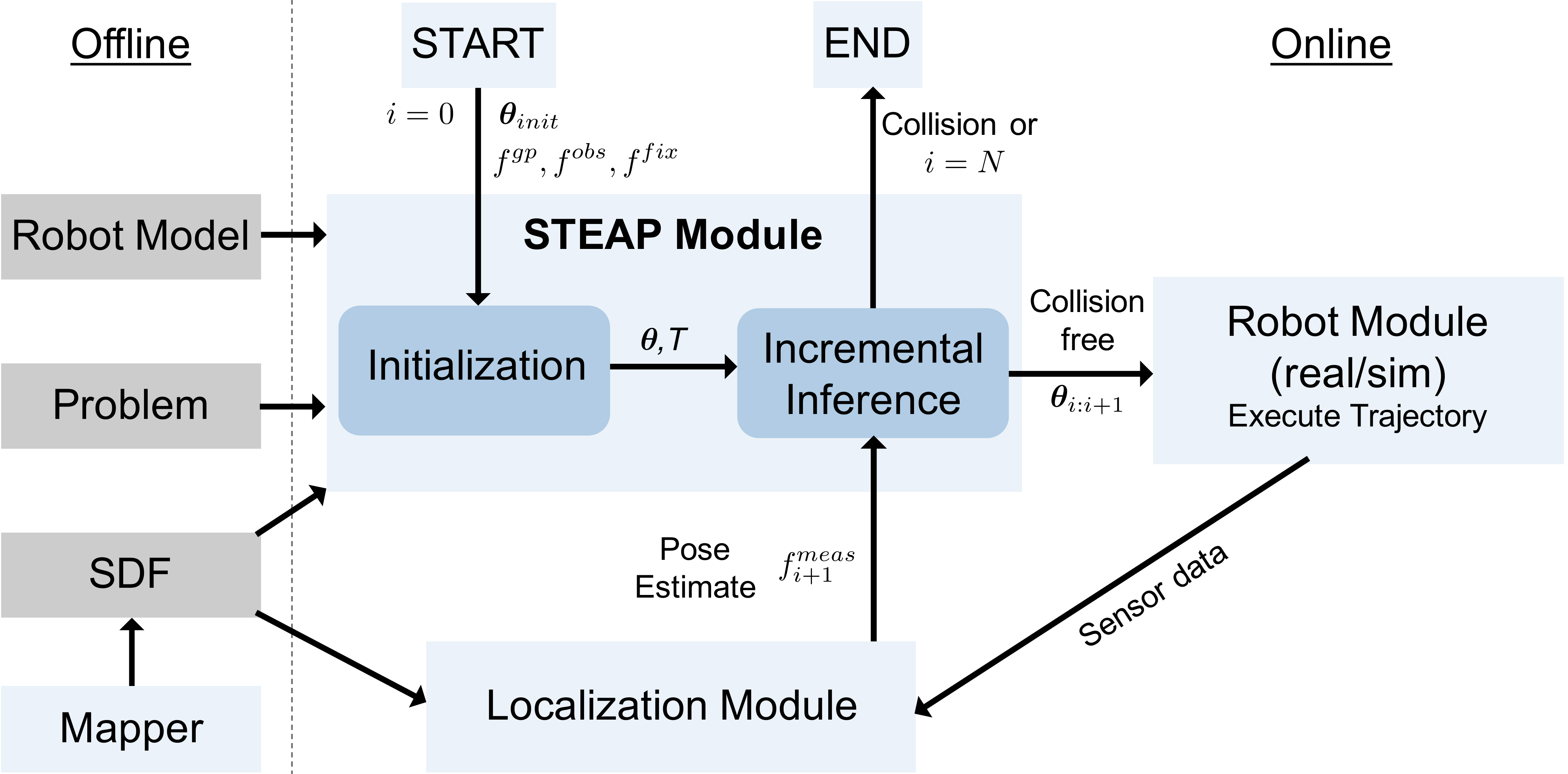}
	\end{minipage}
\end{figure*}

\subsection{GP prior on Lie groups}

The sparse GP prior defined in GPMP2 \citep{Dong-RSS-16} works well for robot manipulators since their configurations can be defined using a \emph{vector space}. But not all robots' configuration spaces can be well represented with a vector space. For example, the orientations of rigid body in 3D space cannot be represented by vectors without singularity (Euler angle) or extra degrees (quaternion). \emph{Lie groups} offer more general robot configuration space representations. For example, SE(2) can represent position and orientation for a planar base of a mobile manipulator, and SE(3) can represent position and orientation for an aerial vehicle. Prior work~\citep{anderson2015full} proposed the sparse GP prior on SE(3) which is useful for trajectory estimation. In this section we extend the sparse GP priors on Lie groups \citep{Chirikjian11book} that will be useful for more general configuration representations. This section is an overview of GP priors on Lie groups, a more detailed discussion can be found in our technical report \citep{dong2017arxiv}.

A $N$-dimensional matrix Lie group $G$ is a sub-group of the general linear group and defines a smooth differentiable manifold whose local tangent space is described by its associated \emph{Lie algebra} $\mathfrak{g}$ \citep{Chirikjian11book}. For example, the Lie algebra of SE(2) is defined by a skew symmetric matrix. We can switch between them using the \emph{exponential map} $\exp : \mathfrak{g} \rightarrow G$ and the \emph{logarithm map} $\log : G \rightarrow \mathfrak{g}$ and to convert elements in local coordinate of $G$ to Lie algebra and vice versa we can use the \emph{hat operator} $\wedge : \mathbb{R}^N \rightarrow \mathfrak{g}$ and the \emph{vee operator} $\vee : \mathfrak{g} \rightarrow \mathbb{R}^N$ respectively \citep{Chirikjian11book}.

\subsubsection{Constant velocity GP prior}

We define a constant velocity GP prior here to match the vector space prior in GPMP2 \citep{Dong-RSS-16}, but just like the vector space, other priors can also be applied here. Let $T \in G$ represent a state in $G$, such that $T(t)$ defines a continuous-time trajectory in $G$. To generate the GP we need to first construct a stochastic differential equation (SDE) with a Markovian state \citep{barfoot2014batch}. Let that state be $\bm{\theta}(t) \doteq \{ T(t), \bm{\varpi}(t) \}$, with the SDE as a double integrator noise-on-acceleration model that will yield a constant velocity prior,
\begin{equation}
\ddot{\bm{\varpi}}(t) = \mathbf{w}(t), \ \ \mathbf{w}(t) \sim GP(\mathbf{0}, \mathbf{Q}_C\delta(t-t')) \label{eq:const_vel_gp}
\end{equation}
where $\bm{\varpi}(t)$ is the `body-frame velocity' variable defined by
\begin{equation}
\bm{\varpi}(t) \doteq (T(t)^{-1} \dot{T}(t))^{\vee}. \label{eq:lie-body-vel}
\end{equation}
and $\mathbf{w}(t)$ is a white noise, zero mean Gaussian process and power-spectral density matrix $\mathbf{Q}_C$ \citep{barfoot2014batch}. Since $\forall \ T \in G, \ T^{-1} \dot{T} \in \mathfrak{g}$~\citep{Chirikjian11book}, we can apply the $\vee$ operator on $T(t)^{-1} \dot{T}(t)$. However, unlike GPMP2 \citep{Dong-RSS-16}, this SDE is non-linear. Fortunately through linearization we can convert it to the linear time-varying SDE (LTV-SDE) that GPMP2 uses.

\subsubsection{Local linearization}

With linearization on the Lie group around any $T_i$, we can define both a \emph{local} GP and a LTV-SDE on the linear tangent space to leverage the constant-velocity GP prior. A local GP at any time t, $t_i \leq t \leq t_{i+1}$ will be
\begin{equation}
T(t) = T_i \exp(\bm{\xi}_i(t)^{\wedge}), \ \ \bm{\xi}_i(t) \sim {N}(\mathbf{0}, \mathbf{{K}}(t_i, t)) \label{eq:local_const_vel_gp}
\end{equation}
where the \emph{local} pose variable $\bm{\xi}_{i}(t) \in \mathbb{R}^N$ 
around $T_i$ is 
\begin{equation}
\bm{\xi}_{i}(t) \doteq \log( T_{i}^{-1} T(t) )^{\vee}.
\end{equation}
Then the local LTV-SDE is defined using the local pose similar to Eq.~\eqref{eq:const_vel_gp} as a double integrator noise-on-acceleration model to give a constant velocity prior
\begin{equation}
\ddot{\bm{\xi}_{i}}(t) = \mathbf{w}(t), \ \ \mathbf{w}(t) \sim GP(\mathbf{0}, \mathbf{Q}_C\delta(t-t')) \label{eq:const_vel_sde}
\end{equation}
with the local Markovian state for the LTV-SDE as 
\begin{equation}
\bm{\gamma}_i(t) \doteq \{ \bm{\xi}_{i}(t), \dot{\bm{\xi}_{i}}(t) \}.
\end{equation}
To show the equivalence between the original nonlinear SDE and the local approximation we use the identity \citep{Chirikjian11book}
\begin{equation}
T(t)^{-1} \dot{T}(t) = \big( \mathbf{{J}}_r(\bm{\xi}_i(t)) \dot{\bm{\xi}}_i(t) \big) ^{\wedge}
\end{equation}
where $\mathbf{{J}}_r$ is the \emph{right Jacobian} of $G$. Following from Eq.~\eqref{eq:lie-body-vel} we get
\begin{equation}
\dot{\bm{\xi}}_i(t) = \mathbf{{J}}_r(\bm{\xi}_i(t))^{-1} \bm{\varpi}(t)
\end{equation}
If the time interval between any $t_i$ and $t_{i+1}$  is small then the approximation
\begin{equation}
\dot{\bm{\xi}}_i(t) \approx  \bm{\varpi}(t)
\end{equation}
is good. Then linear SDE Eq.~\eqref{eq:const_vel_sde} is a good approximation of non-linear SDF Eq.~\eqref{eq:const_vel_gp}, so we linearize the SDE and we can apply LTV-SDE GP prior by~\citep{barfoot2014batch}.

\subsubsection{GP prior factor}
 We define the GP factor between any two states at $t_i$ and $t_{i+1}$ with their local pose using the linearized SDE formulation on the Lie group manifold
\begin{align}\label{eq:gp_prior_factor}
f&^{gp}_i(\bm{\theta}_i, \bm{\theta}_{i+1}) \nonumber \\ 
&= \exp \bigg\{\hspace{-1mm} - \frac{1}{2} \| \bm{\Phi}(t_{i+1}, t_{i})\bm{\gamma}_{i} - \bm{\gamma}_{i+1} \|^{2}_{\mathbf{Q}_{i,i+1}} \hspace{-1mm} \bigg\}
\end{align}
where the logarithm map provide transformation from SE(2) to $\mathbb{R}^3$ and vice versa with the exponential map, and 
\begin{equation}
\bm{\Phi}(t, s) = \begin{bmatrix} \mathbf{1} & (t-s) \mathbf{1} \\
\mathbf{0} & \mathbf{1} \end{bmatrix}, 
\mathbf{Q}_{i,i+1} = \begin{bmatrix} \frac{1}{3} \Delta t_i^3 \mathbf{Q}_C &
\frac{1}{2} \Delta t_i^2 \mathbf{Q}_C \\ 
\frac{1}{2} \Delta t_i^2 \mathbf{Q}_C &
\Delta t_i \mathbf{Q}_C \end{bmatrix}.
\end{equation}
We use this GP prior in our current implementation.


\section{Implementation Details}\label{sec:imp}

We implement STEAP within the PIPER \citep{mukadam2017piper} package using ROS \citep{ros2009} and GPMP2 \citep{Dong-RSS-16} and have open-sourced the code. Fig.~\ref{fig:block} shows a block diagram of the framework. The offline phase assimilates (i) robot-specific information including model and physical parameters, (ii) problem definitions and optimization parameters, and (iii) a pre-generated signed distance field (SDF) of the environment, which is assumed to be static. In the online phase, this information is passed to our central module, \texttt{STEAP Module}, that solves \algo problems and communicates with the \texttt{Robot Module} (simulated or physical) with sensors, and the \texttt{Localization Module} that takes raw sensor measurements and outputs a noisy pose estimate for the robot that can be interpreted by the \texttt{STEAP Module}.

Note that in our framework, the \texttt{Localization Module} is free to be any source of raw or processed sensor information, as long as suitable factors are defined to fuse sensor information in the factor graph, e.g. GPS, LIDAR, monocular, or stereo camera data. In our implementation we use a depth image-based localization algorithm, detailed in Section~\ref{sec:icp}.

\subsection{STEAP module}\label{sec:interface}

This module uses the robot and problem configuration. In the first step, the module initializes the problem by constructing an initial factor graph, performing inference to get the first planned solution, and constructing a Bayes tree for future iterations. During every other step, the module performs incremental inference by updating the Bayes tree directly. At every step, the replanned solution is upsampled and checked for safety, and is sent to the \texttt{Robot Module}. When the \texttt{Localization Module} returns a current pose measurement, new sensor measurement factors are added to the Bayes tree and the updated posterior is evaluated. This procedure repeats until the full trajectory completes execution or exits due to collision failure. Given the generic implementation of this module, our framework can be used on any simulated or real robot as long as the robot information is provided in the offline phase.

\subsection{Robot module}\label{sec:robot}

This module consists of the robot API and controllers that can understand and execute the trajectory passed to it by the \texttt{STEAP Module} for either a real robot or an interface with Gazebo for a simulated robot. Sensors on the robot pass information to the \texttt{Localization Module}. Note that in our simulator, adjustable Gaussian noise is mixed in both system dynamics (more precisely vehicle velocities) and sensor measurements (more precisely depth measurements of depth camera simulator), to simulate the real-world stochasticity. 

\subsection{Mapper}\label{sec:mapper}

\begin{figure}[t]
	\centering
	\includegraphics[width=0.338\linewidth]{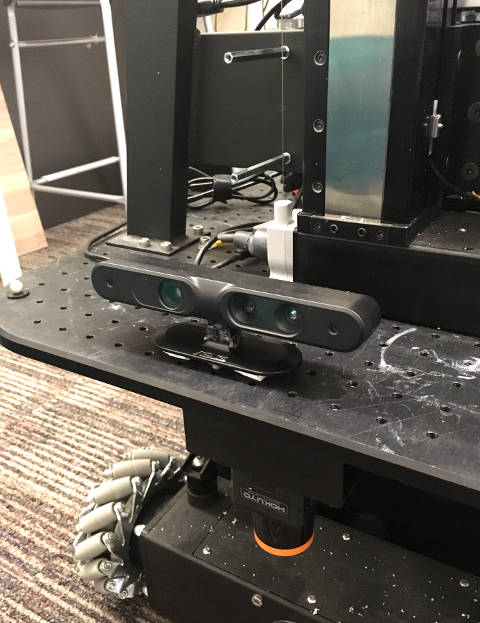}
	\includegraphics[width=0.64\linewidth]{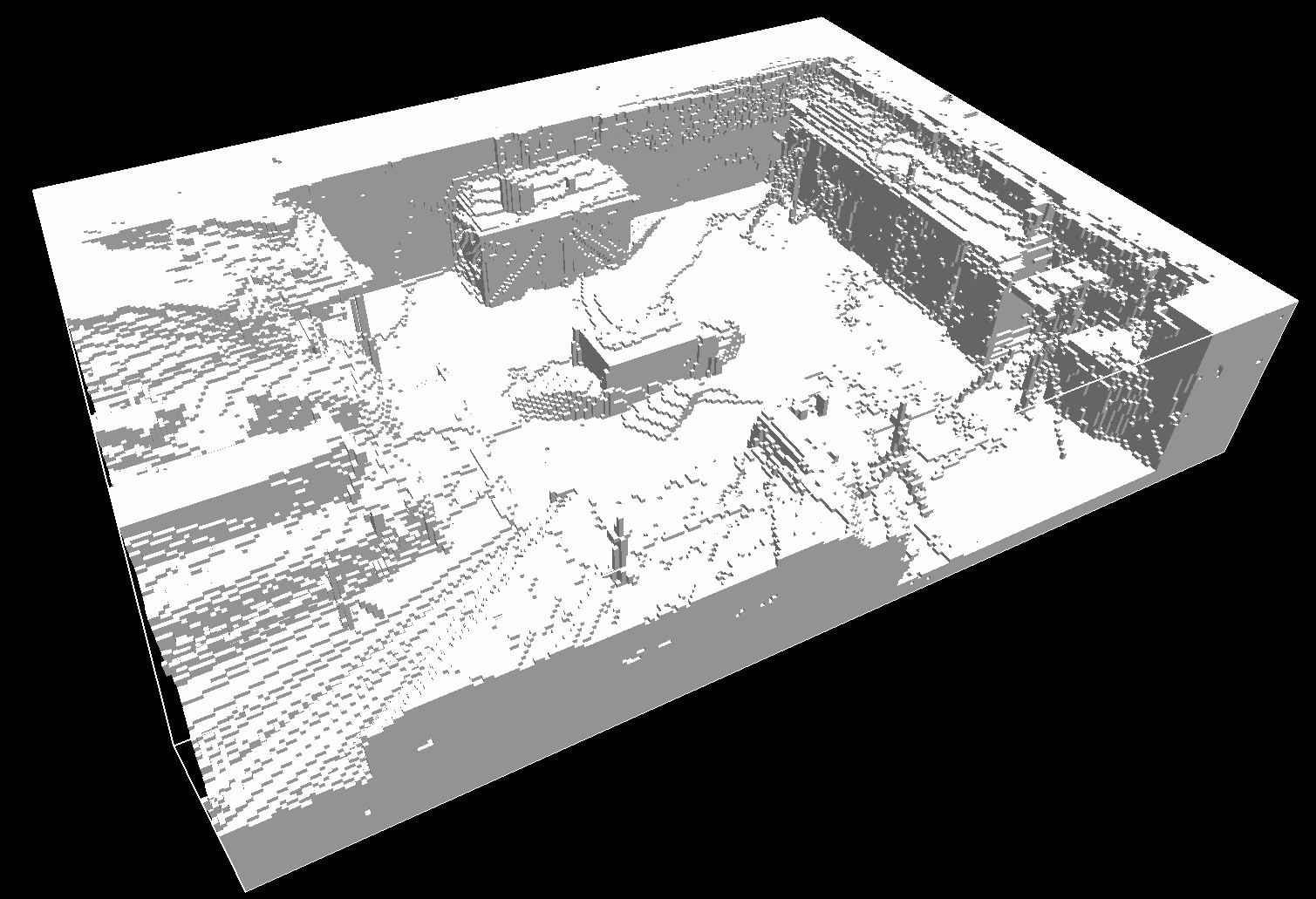}  
	\caption{Left: the PrimeSense depth camera mounted on the robot base. Right: one 8m$\times$6m$\times$1.5m occupancy grid map built by the mapper module.}
	\label{fig:map}
\end{figure}

Obstacle factors require a signed distance field (SDF) to calculate obstacle cost. Although we can calculate SDFs from CAD models in simulation, we will not have these models available in most real-world environments. Therefore, we build SDFs from sensor data. We use depth scans from depth sensors (a simulated depth camera in simulation and a PrimeSense, shown in Fig.~\ref{fig:map}, in real-world experiments), and an occupancy grid mapping approach \citep[Ch.9]{thrun2005probabilistic} to generate the SDF. The space is first discretized into small cells, and a probability of occupancy $p_{o}$ is assigned to each cell, with initial $p_{o} = 0.5$ (since we have no information). All $p_{o}$s are updated by sensor measurements, and after all sensor data is received, we assume cells with $p_{o} \geq 0.5$ are occupied, and cells with $p_{o} < 0.5$ are unoccupied. Note that we assume cells with $p_{o} = 0.5$, which indicates no depth measurement available, are occupied, since it is safer to assume that locations never observed are occupied by obstacles. True camera poses are provided by a motion capture system during the mapping process. After occupancy grid mapping, we calculate the signed distance field by the efficient distance transformation \citep{felzenszwalb2004distance}. We use CUDA~\citep{nickolls2008cuda} to implement occupancy grid mapping and distance transformation, allowing us to achieve real-time performance on scenes roughly of size 8m$\times$6m$\times$1.5m with 3cm resolution. A map constructed with this approach is shown in Fig.~\ref{fig:map}.

\subsection{Localization module}\label{sec:icp}

The Localization module reads raw sensor data from the \texttt{Robot Module}, calculates a pose estimate of the robot, and provides this information to the \texttt{STEAP Module}. Here it's free to choose any Localization algorithm. We choose an ICP-style iterative approach, which similar to tracking in KinectFusion \citep{izadi2011kinectfusion} but operates on the full SDF generated by the \texttt{Mapper} rather than the truncated SDF (TSDF) in KinectFusion. We use CUDA to implement and parallelize the tracking module to achieve real-time performance. Although additional sensor data like RGB images, odometry, and laser scans are available to the robot, we only use depth images from PrimeSense in our experiments.


\subsection{Computational complexity}\label{sec:comp}

Since our approach can involve performing inference on arbitrary factor graphs and incremental inference on their corresponding Bayes trees, the computational complexity depends on the problem and the structure of the graph or tree and how the tree changes over time. Considering the case of the experiments in this paper where the graphs are chain-like and measurement factors are unary, the complexity of the batch step is $O(TD^2)$, where $T$ is the number of time-steps (or states) and $D$ is the dimension of the system state. The time complexity $O(TD^2)$ comes from solving the block-tridiagonal linear system with block size $D \times D$.

In the online step, the time complexity of the incremental update is $O(VD^2)$, where $V$ is the total number of variables (or states) in all the affected cliques of the Bayes tree starting from the root of the tree. In the beginning since all states in the Bayes tree are affected, $V$ is the same as $T$ and the time complexity is equal to the batch step complexity. When the robot is moving along the trajectory, $V$ is much smaller, as shown in Fig.~\ref{fig:steap_bayes_tree}.

The time complexity for STEAP discussed above only applies to cases of chain-like graphs (as shown in Fig.~\ref{fig:example}) that have a start-to-goal variable ordering. In the presence of non-unary measurement factors (for example, higher-order measurement factors like landmark observations) that connect to states across greater time-intervals, or in the case of loop-closures, the time complexity will be greater. In general, the time complexity increases as the sparsity in the graph decreases, and is also dependent on the variable ordering selected by the COLAMD heuristics.


\section{Evaluation}\label{sec:eval}

\begin{figure}[!t]
	\centering
	\includegraphics[trim={3.3cm 0 2.8cm 0},clip,width=0.6\linewidth]{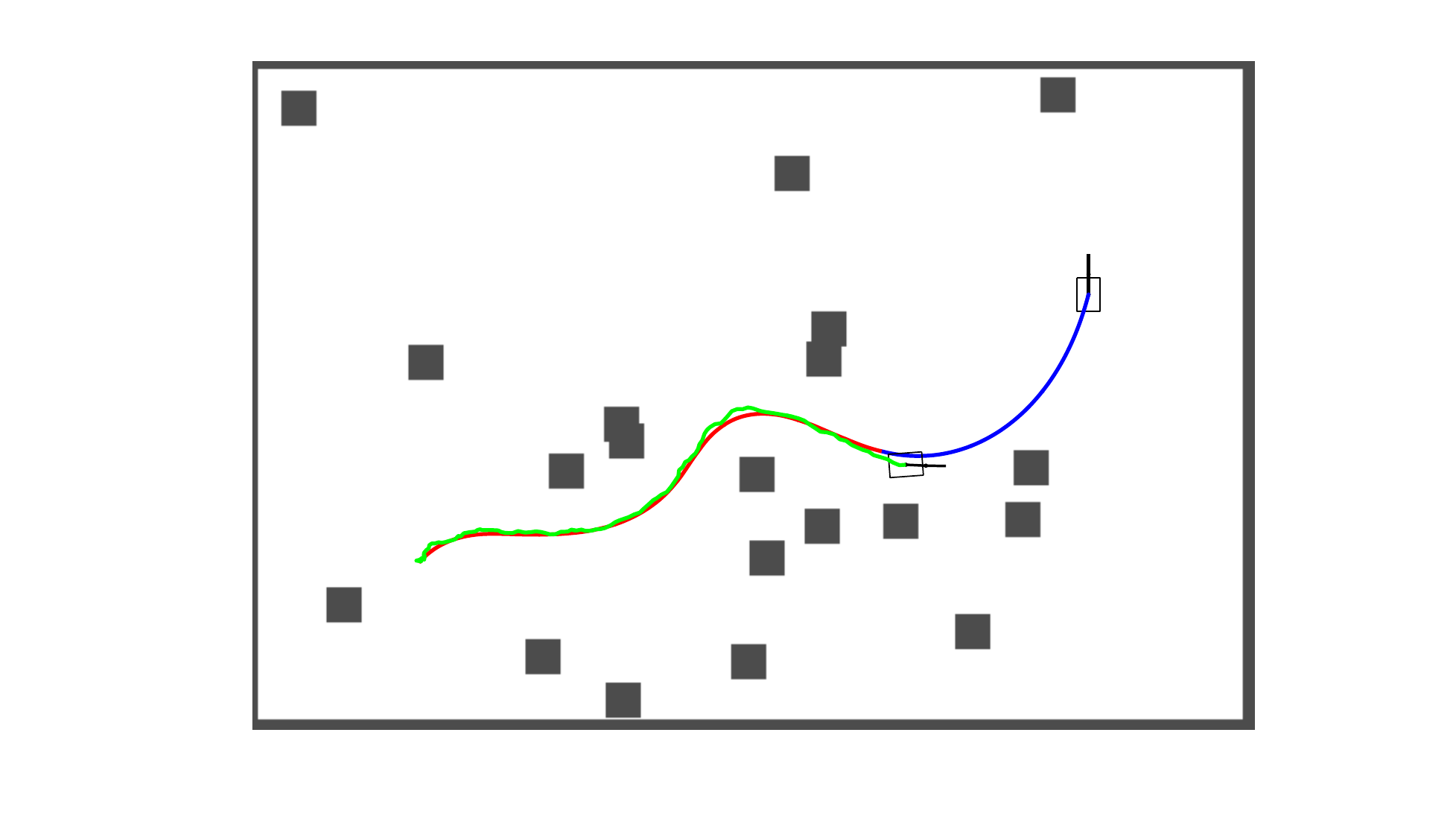}
	\caption{\algo result on an example from the simulation benchmark with a planar 2-link mobile arm. Ground-truth (green), estimated past (red) and replanned future (blue) trajectories are shown with the current robot pose between the red and blue trajectories and the goal at the end of the blue trajectory. Best viewed in color.}
	\label{fig:2dmarm}
\end{figure}

We evaluate our framework with simulation and real-world experiments.\footnote{A video of experiments is available at \url{https://youtu.be/lyayNKV1eAQ}\label{fn:exp}} The simulation benchmark is performed on two datasets: a planar 2-link mobile arm in several randomly generated 2D environments as shown in Fig.~\ref{fig:2dmarm}, and a 18-DOF PR2 robot in a simulated indoor environment as shown in Fig.~\ref{fig:pr2simu}. The real-world experiments are performed on a mobile manipulator, with an omni-drive base and a 6-DOF Kinova JACO2 arm, in an indoor environment as shown in Fig. \ref{fig:vector}.

In our experiments we compare our proposed approach, simultaneous trajectory estimation and planning (\texttt{\algo}), against an open loop execution (\texttt{\ol}). In \texttt{\ol} the initial inference solves the GPMP2 planning problem and the planned trajectory is executed without any estimation or replanning until the final time-step or a collision. In our simulation experiments, we also compared against simultaneous localization and planning (\texttt{\slap}) that uses the current measurement factor in the graph, but updates only a truncated version of the graph associated with the future states to replan. This graph also uses the GP prior factors from \texttt{\algo}.


\subsection{Benchmark with a planar 2-link mobile arm}\label{sec:sim}

For this benchmark we use a simulated 2-link planar mobile arm with base of size $1m \times 0.7m$ and link length $0.6m$ in an environment of size $30m \times 20m$. The robot is equipped with a simulated 2D laser scanner for localization using ICP. The environment is populated with 20 randomly generated obstacles of size $1m \times 1m$. The graph consists of 30 states from start to goal with 5 interpolated binary obstacle factors between any two states. We compare \texttt{\ol}, \texttt{\slap} and \texttt{\algo} across different amounts of robot dynamics noise ($n_{dyn}$), implemented as uniform bounded additive noise ($m/s$) to the robot velocity, and camera measurement noise ($n_{cam}$), implemented as additive Gaussian noise ($m^{-1}$) when receiving depth information from the camera on the robot. Each setting is run with 40 distinct seeds (each seed yields a new environment) to account for stochasticity, which are kept the same across all three scenarios. In each trial we record if the trajectory successfully finishes without collision (success), the distance from the goal (goal error) at the end of execution, and $L_2$ norm of the ground-truth trajectory with the estimated trajectory (estimation error).

\begin{table}[t]
\begin{center}
\begin{subtable}{\linewidth}
\caption{Success rate on planar 2-link mobile arm benchmark.}
\label{table:succ}
\centering
\begin{tabular}{|c|c|c|cc|cc|}
\hline
 & $\mathbf{n_{cam}}$ & & \multicolumn{2}{c|}{\bf{0.02}} & \multicolumn{2}{c|}{\bf{0.1}}\\
 & & \texttt{\ol} & \texttt{\slap} & \texttt{\algo} & \texttt{\slap} & \texttt{\algo}\\ 
\hline
\multirow{3}{*}{$\mathbf{n_{dyn}}$}
& \bf{0.1} & 0.5250  &  0.1250  &  0.5750  &  0.0750  &  0.7000 \\
& \bf{0.2} & 0.3250   & 0.1250  &  0.6000 &   0.1000  &  0.4750 \\
& \bf{0.5} & 0.2500 &   0.0750  &  0.3750  &  0.0500  &  0.4500 \\
\hline 
\end{tabular}
\end{subtable}

\par\bigskip

\begin{subtable}{\linewidth}
\caption{Goal translational error (in $m$).}
\label{table:goal}
\centering
\begin{tabular}{|c|c|c|cc|cc|}
\hline
 & $\mathbf{n_{cam}}$ & & \multicolumn{2}{c|}{\bf{0.02}} & \multicolumn{2}{c|}{\bf{0.1}} \\
 & & \texttt{\ol} & \texttt{\slap} & \texttt{\algo} & \texttt{\slap} & \texttt{\algo} \\ 
\hline
\multirow{3}{*}{$\mathbf{n_{dyn}}$}
& \bf{0.1} & 0.7463  &  0.7556 &  0.3413   & 0.7067  &  0.4377\\
& \bf{0.2} & 0.9715  &  0.8012  &  0.4888 &   1.0793 &   0.5162\\
& \bf{0.5} & 1.4179  &  1.2110   & 0.6872  &  1.1295 &   0.7345\\
\hline 
\end{tabular}
\end{subtable}

\par\bigskip

\begin{subtable}{\linewidth}
\caption{Goal rotational error (in $rad$).}
\label{table:goal_rot}
\centering
\begin{tabular}{|c|c|c|cc|cc|}
\hline
 & $\mathbf{n_{cam}}$ & & \multicolumn{2}{c|}{\bf{0.02}} & \multicolumn{2}{c|}{\bf{0.1}} \\
 & & \texttt{\ol} & \texttt{\slap} & \texttt{\algo} & \texttt{\slap} & \texttt{\algo} \\ 
\hline
\multirow{3}{*}{$\mathbf{n_{dyn}}$}
& \bf{0.1} & 0.0805 &   0.1978  &  0.0269 &   0.0497 &   0.0379\\
& \bf{0.2} & 0.0952 &   0.2344  &  0.0433 &   0.0847 &   0.0401\\
& \bf{0.5} & 0.0952  &  0.2344   & 0.0433  &  0.0847  &  0.0401\\
\hline 
\end{tabular}
\end{subtable}

\par\bigskip

\begin{subtable}{\linewidth}
\caption{Estimation translational error (in $m$).}
\label{table:loc}
\centering
\begin{tabular}{|c|c|cc|cc|}
\hline
 & $\mathbf{n_{cam}}$ & \multicolumn{2}{c|}{\bf{0.02}} & \multicolumn{2}{c|}{\bf{0.1}} \\
 & & \texttt{\slap} & \texttt{\algo} & \texttt{\slap} & \texttt{\algo}\\ 
\hline
\multirow{3}{*}{$\mathbf{n_{dyn}}$}
& \bf{0.1} & 0.3662 &   0.1598 &   0.9217  &  0.2213 \\
& \bf{0.2} & 0.3644 &   0.2183 &   0.8885 &   0.2242 \\
& \bf{0.5} & 0.3937   & 0.3065  &  1.0125  &  0.3309 \\
\hline 
\end{tabular}
\end{subtable}

\par\bigskip

\begin{subtable}{\linewidth}
\caption{Estimation rotational error (in $rad$).}
\label{table:loc_rot}
\centering
\begin{tabular}{|c|c|cc|cc|}
\hline
 & $\mathbf{n_{cam}}$ & \multicolumn{2}{c|}{\bf{0.02}} & \multicolumn{2}{c|}{\bf{0.1}} \\
 & & \texttt{\slap} & \texttt{\algo} & \texttt{\slap} & \texttt{\algo}\\ 
\hline
\multirow{3}{*}{$\mathbf{n_{dyn}}$}
& \bf{0.1} & 0.0470 &   0.0183  &  0.1287 &   0.0266 \\
& \bf{0.2} & 0.0480 &   0.0280  &  0.1132 &   0.0297 \\
& \bf{0.5} & 0.0527  &  0.0450   & 0.1265  &  0.0468 \\
\hline 
\end{tabular}
\end{subtable}

\end{center}
\end{table}

The results for this benchmark are summarized in Table \ref{table:succ}--\ref{table:loc_rot}. The goal and estimation error are aggregates of runs where success is true (the robot reached the goal without colliding with any obstacles). As expected, \texttt{\ol} exhibits a low success rate that drops further with an increase in $n_{dyn}$. \texttt{\slap} has lower success rates compared to \texttt{\ol}; a possible reason is the low precision trajectory estimation from \texttt{\slap} confuses the robot. Comparatively \texttt{\algo} has a higher success rate than \texttt{\slap} and \texttt{\ol}, and also follows the decreasing trend with increasing $n_{dyn}$.

The goal error in \texttt{\algo} is much lower than \texttt{\slap} and \texttt{\ol}, with the help from high precision trajectory estimation. The trajectory estimation error is significantly smaller with \texttt{\algo} compared to \texttt{\slap} and the difference between them increases with increasing $n_{cam}$. This can be attributed to simultaneously solving the trajectory estimation and planning problems; the motion plan helps in providing a better estimate of the robot's trajectory and in turn a better estimate of the trajectory helps in generating a better motion plan.

\subsection{Benchmark with 18-DOF full-body PR2}\label{sec:sim_pr2}

\begin{figure}[!t]
	\centering
	\includegraphics[width=0.98\linewidth]{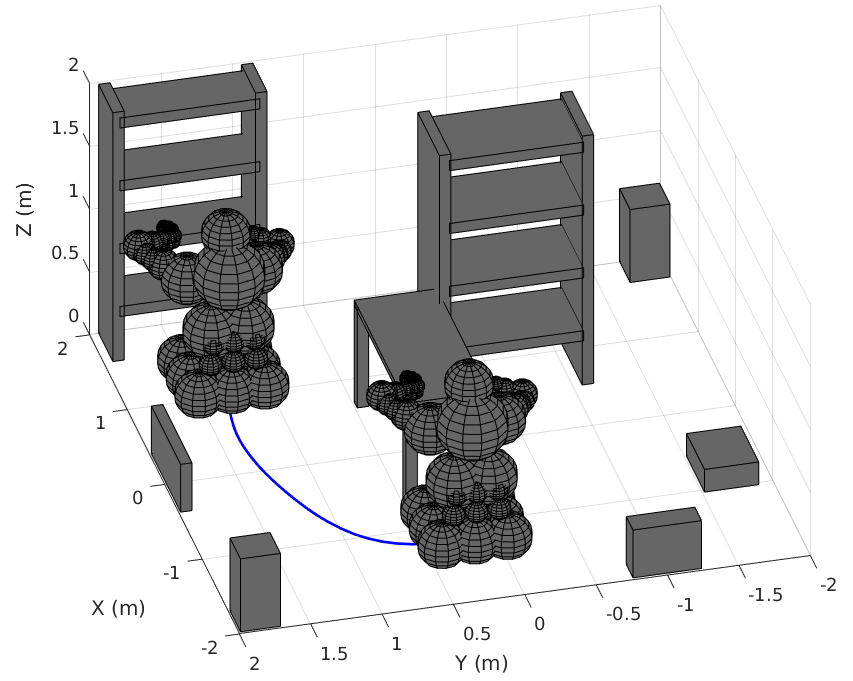}
	\caption{Workspace setup for PR2 full-body problems. An example problem with start and end robot state, and planned trajectory of the PR2 base is shown. As done in GPMP2, the PR2 body is approximated with a set of spheres for collision checking.}
	\label{fig:pr2simu}
\end{figure}

We additionally evaluated our approach on a simulation dataset in 3D, with a 18-DOF PR2 robot in an indoor setting. The PR2 has two 7-DOF arms (without grippers), a 3-DOF mobile base, 1-DOF linear actuator that moves the torso in the vertical direction, and a 2D laser scanner on the robot base (10cm above the ground level) used for localization. The environment consists of two cabinets, a desk and several random obstacles so that the laser scanner receives non-trivial readings when the robot is facing outwards. The benchmark consists of a total of 6 tasks, each with a start and a goal state, and the robot is tasked with reaching the goal states from the start states. The environment and an example problem from the benchmark is shown in Fig.~\ref{fig:pr2simu}. We compare \texttt{\ol}, \texttt{\slap} and \texttt{\algo} across different $n_{dyn}$ and $n_{cam}$ settings, with a graph of 40 states from start to goal. Each setting is run for 120 trials (6 tasks with 20 distinct seeds) and similar metrics as the 2D benchmarks are recorded.

\begin{table}[t]
\begin{center}
\begin{subtable}{\linewidth}
\caption{Success rate on 18-DOF full-body PR2 benchmark.}
\label{table:succ_pr2}
\centering
\begin{tabular}{|c|c|c|cc|cc|}
\hline
 & $\mathbf{n_{cam}}$ & & \multicolumn{2}{c|}{\bf{0.1}} & \multicolumn{2}{c|}{\bf{0.5}} \\
 & & \texttt{\ol} & \texttt{\slap} & \texttt{\algo} & \texttt{\slap} & \texttt{\algo} \\  
\hline
\multirow{3}{*}{$\mathbf{n_{dyn}}$}
& \bf{0.02} & 0.3833 &   0.7500 &  0.8333  &  0.5417   & 0.6917 \\
& \bf{0.1} & 0.0750  &  0.3083 &   0.8583 &   0.0667  &  0.6250 \\
& \bf{0.5} & 0.0083  &  0.0500 &   0.3500 &   0.0083 &   0.3083 \\
\hline 
\end{tabular}
\end{subtable}

\par\bigskip

\begin{subtable}{\linewidth}
\caption{Goal translational error (in $m$).}
\label{table:goal}
\centering
\begin{tabular}{|c|c|c|cc|cc|}
\hline
 & $\mathbf{n_{cam}}$ & & \multicolumn{2}{c|}{\bf{0.1}} & \multicolumn{2}{c|}{\bf{0.5}} \\
 & & \texttt{\ol} & \texttt{\slap} & \texttt{\algo} & \texttt{\slap} & \texttt{\algo} \\ 
\hline
\multirow{3}{*}{$\mathbf{n_{dyn}}$}
& \bf{0.02} & 0.2032 &   0.0798  &  0.0426 &   0.0970 &   0.0943 \\
& \bf{0.1} & 0.4197 &   0.1108  &  0.0440  &  0.1647  &  0.1024 \\
& \bf{0.5} & 2.8040  &  0.1357   & 0.0936   & 0.4100   & 0.1238 \\
\hline 
\end{tabular}
\end{subtable}

\par\bigskip

\begin{subtable}{\linewidth}
\caption{Goal rotational error (in $rad$).}
\label{table:goal}
\centering
\begin{tabular}{|c|c|c|cc|cc|}
\hline
 & $\mathbf{n_{cam}}$ & & \multicolumn{2}{c|}{\bf{0.1}} & \multicolumn{2}{c|}{\bf{0.5}} \\
 & & \texttt{\ol} & \texttt{\slap} & \texttt{\algo} & \texttt{\slap} & \texttt{\algo} \\ 
\hline
\multirow{3}{*}{$\mathbf{n_{dyn}}$}
& \bf{0.02} & 0.1078 &   0.0396 &   0.0219 &   0.0587 &   0.0543\\
& \bf{0.1} & 0.3206  &  0.0690  &  0.0261 &   0.0944 &   0.0644\\
& \bf{0.5} & 0.5170   & 0.0803  &  0.0473  & 0.2780 &   0.0784\\
\hline 
\end{tabular}
\end{subtable}

\par\bigskip

\begin{subtable}{\linewidth}
\caption{Estimation translational error (in $m$).}
\label{table:loc}
\centering
\begin{tabular}{|c|c|cc|cc|}
\hline
 & $\mathbf{n_{cam}}$ & \multicolumn{2}{c|}{\bf{0.1}} & \multicolumn{2}{c|}{\bf{0.5}} \\
 & & \texttt{\slap} & \texttt{\algo} & \texttt{\slap} & \texttt{\algo}\\ 
\hline
\multirow{3}{*}{$\mathbf{n_{dyn}}$}
& \bf{0.02} & 0.0807  &  0.0287  &  0.1487  &  0.0739 \\
& \bf{0.1} & 0.1777  &  0.0307 &   0.1749  &  0.0728 \\
& \bf{0.5} & 0.2100  &  0.0595  &  0.2550  &  0.0858 \\
\hline 
\end{tabular}
\end{subtable}

\par\bigskip

\begin{subtable}{\linewidth}
\caption{Estimation rotational error (in $rad$).}
\label{table:loc}
\centering
\begin{tabular}{|c|c|cc|cc|}
\hline
 & $\mathbf{n_{cam}}$ & \multicolumn{2}{c|}{\bf{0.1}} & \multicolumn{2}{c|}{\bf{0.5}} \\
 & & \texttt{\slap} & \texttt{\algo} & \texttt{\slap} & \texttt{\algo}\\ 
\hline
\multirow{3}{*}{$\mathbf{n_{dyn}}$}
& \bf{0.02} & 0.0415 &   0.0168 &   0.0689 &   0.0423 \\
& \bf{0.1} & 0.0906 &   0.0185 &   0.0790  &  0.0459 \\
& \bf{0.5} & 0.1007  &  0.0308  &  0.1310  &  0.0475 \\
\hline 
\end{tabular}
\end{subtable}

\par\bigskip

\begin{subtable}{\linewidth}
\caption{Average runtime (in $s$).}
\label{table:time_pr2}
\centering
\begin{tabular}{|c|c|cc|cc|}
\hline
 & $\mathbf{n_{cam}}$ & \multicolumn{2}{c|}{\bf{0.1}} & \multicolumn{2}{c|}{\bf{0.5}} \\
 & & \texttt{\slap} & \texttt{\algo} & \texttt{\slap} & \texttt{\algo}\\ 
\hline
\multirow{3}{*}{$\mathbf{n_{dyn}}$}
& \bf{0.02} & 0.1181  &  0.0141  &  0.1273  &  0.0188 \\
& \bf{0.1} & 0.1377  &  0.0149  &  0.1353  &  0.0193 \\
& \bf{0.5} & 0.1474  &  0.0185  &  0.1438 &   0.0195 \\
\hline 
\end{tabular}
\end{subtable}

\end{center}
\end{table}

The results for this benchmark are summarized in Table~\ref{table:succ_pr2}-\ref{table:time_pr2}, which support the finding in the previous 2D benchmark.
We see that \texttt{\algo} has higher success rates compared to \texttt{\slap} and \texttt{\ol}. This is a result of the lower estimation errors (compared to \texttt{\slap}) and better quality plans achieved through joint inference. \texttt{\algo} has a significantly lower estimation errors with larger $n_{cam}$. The goal error in \texttt{\slap} and \texttt{\algo} are smaller than \texttt{\ol} since both are feedback approaches that can reduce the error when approaching the goal. We also observe that \texttt{\algo} has a smaller performance drops caused by a high value of both $n_{dyn}$ and $n_{cam}$, than \texttt{\slap} and \texttt{\ol}.

We also report timing results for this benchmark. On average \texttt{\algo} takes about 17ms per time-step, which is significantly faster than SLAP with an average of about 130ms per time-step.


\begin{figure}[!t]
	\centering
	\includegraphics[width=0.98\linewidth]{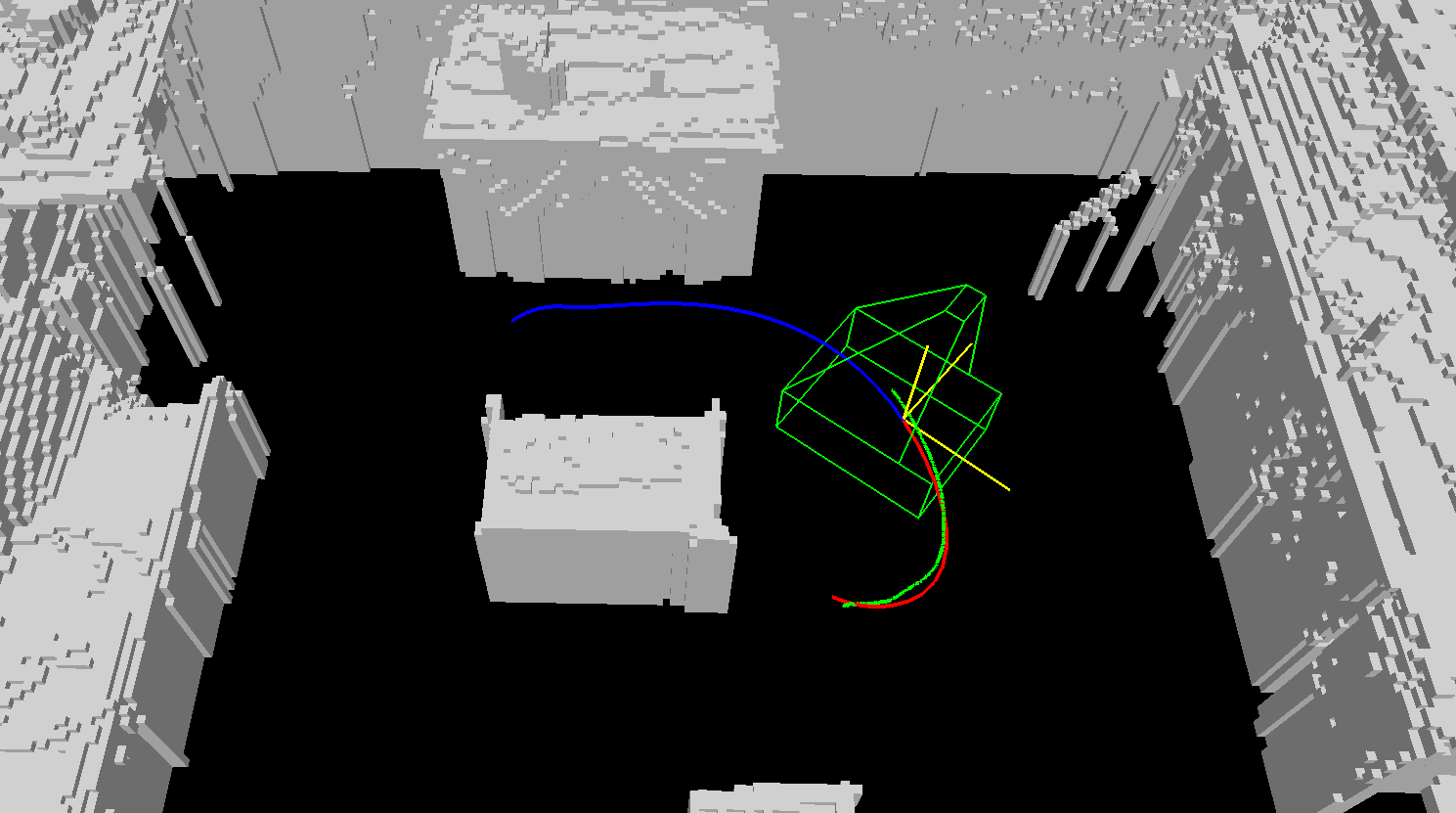}  
	\caption{Visualization of STEAP results on one run of problem 1. The green line is the ground-truth trajectory as determined by the motion capture system, and green robot outline shows the current pose of the Vector robot. The blue line is the planned trajectory and the red line is the estimated trajectory. The yellow axis is current raw pose estimate. The ground plane is cut for visibility. Best viewed in color.}
	\label{fig:steap_vector}
\end{figure}

\subsection{Experiments with a real robot}\label{sec:real}

Real-world experiments were performed in an $8m\times6m$ indoor environment. Various obstacles (desks, sofas and small objects like boxes and cans) are placed in the environment, to simulate domestic scenes. During the experiments, ground-truth robot trajectories are recorded by an Optitrack motion capture system. A photo of the robot traversing the environment and a map of the environment can be found in Fig.~\ref{fig:vector} and Fig.~\ref{fig:map}, respectively.
Our implementation runs on a desktop computer equipped with Intel 4.0GHz quad-core CPU, 32GB memory and one NVIDIA Titan X GPU. Robot sensor data is streamed to the desktop over WiFi, and STEAP commands are streamed back to robot after processing. 

We design 2 problems for performance evaluation. In each problem the robot begins from a start configuration and is tasked with reaching the goal configuration. For both problems the graph consists of 50 states from start to goal with 2 interpolated binary obstacle factors between any two states. Fig.~\ref{fig:steap_vector} shows a screenshot when  the robot is running STEAP for problem 1.
To evaluate the performance of our STEAP implementation, we performed 10 runs for each problem, in which 5 runs switch the start and goal configurations. We record the planned, estimated and ground-truth trajectories and calculate the same performance criteria as in simulation: success rate, final goal error, and trajectory estimation error.

\begin{table}[t]
	\begin{center}
		\caption{Real-world experimental results}
		\label{table:succ_realexp}
		\centering
		\begin{tabular}{|c|cc|}
			\hline
			& \bf{Problem 1} & \bf{Problem 2}\\
			\hline
			\texttt{\ol} success rate & 0/10 & 0/10\\
      \hdashline
			\texttt{\algo} success rate & 9/10 & 10/10\\
			Goal translation error (cm) & 14.20 & 5.19\\
			Localization error (cm) & 7.07 & 6.45\\
			Trajectory estimation error (cm) & 3.48 & 2.53\\
			\hline
		\end{tabular}
	\end{center}
\end{table}

Table.~\ref{table:succ_realexp} reports performance in these real-world experiments. We first run one-time batch planning by GPMP2 and use an open loop controller to follow the planned trajectory. Since the control command execution on the omni-directional wheels is noisy, the robot base cannot follow the planned trajectory well,  so every run ends with a collision.
With the state estimation and replanning provided by STEAP, the robot can follow planned trajectories better, and compensate for perturbations. With STEAP the robot can achieve a 95\% overall success rate for the given tasks, with a final translation error of about 14.2cm in problem 1, and 5.19cm in problem 2. This goal error is due to the finite-horizon trajectory setup that we use. Since if the robot overshoots when near the end of the trajectory, it may not have enough time steps left to recover. The goal error can be reduced with a receding-horizon formulation of our problem.

In addition to improving planning results, STEAP helps with trajectory estimation. We show the raw localization error in Table.~\ref{table:succ_realexp}. Due to the noisy depth measurements, the localization module provides poor estimates of the robot pose. Sometimes the localization module additionally fails due to the scene being out of  sensor range (for example when the robot is too close to obstacles). With STEAP, we can reduce the estimation error by about 50-60\% as shown in Table.~\ref{table:succ_realexp}. In the video of experiments,\textsuperscript{\ref{fn:exp}} one can see that although the raw localization positions have significant jumps between each measurement, the estimation results in STEAP are stabilized given previous sensor information and the planned trajectory.

To evaluate the efficiency of our implementation, we time the localization and STEAP modules separately. Timing results show that, in real-world experiments, localization and STEAP modules have average runtimes of 19.3ms and 76.0ms respectively, and maximum runtimes of 30.3ms and 149ms respectively, indicating that our localization implementation can easily process the depth image stream at 30Hz, and run STEAP at $\sim$ 10Hz.


\section{Limitations and Future Work}\label{sec:diss}

The primary limitation of our current work is that it is only applicable in known, static environments. We use an existing map of the environment and precompute a signed distance field for collision checking. In dynamic environments, the map and the signed distance field would need to be constantly updated, which can be a major computational bottle neck, especially in large environments. Using techniques like incremental mapping \citep{Yan17ras}, incremental signed distance fields \citep{oleynikova2017voxblox}, and dynamic tracking \citep{schmidt2014dart}, we can extend our method to perform SLAM and planning simultaneously online and handle dynamic environments.

Our current implementation is limited to holonomic systems, like the omni-directional mobile base we use in our experiments. It does not support nonholonomic and inequality constraints. But, they can be incorporated as soft constraints with small covariances, for example, the configuration and velocity limit factors~\citep{mukadam2017continuous}. A sequential quadratic programming type procedure would have to be set up to handle hard constraints.

Our approach is a local trajectory optimization method and is, therefore, prone to local minima. In the context of estimation, trajectory optimization rarely suffers from bad local minimas, since obtaining reasonable initial values is not hard during estimation (e.g. from odometry). In the context of planning, trajectory optimization suffers from bad local minimas (ones that are in collision) primarily in extremely cluttered or maze type environments due to the nature of the optimization methods used. Readers are encouraged to refer to \citet{Dong-RSS-16} and \citet{mukadam2017continuous} for some quantitative evaluations on several commonly used trajectory optimization techniques. In case of batch optimization i.e. the first inference step, several ideas like random initializations and graph-based initializations \citep{Huang-ICRA-17} exist to improve results. However, there are no known methods to address this problem during incremental inference using the Bayes tree, except by re-solving a new batch optimization problem, which will be computationally expensive. Developing new incremental algorithms that are better able to contend with local minima is an interesting research direction.

Our approach is capable of fusing information from multiple sensors in an asynchronous fashion using GP interpolation~\citep{Yan17ras}, we are interested in exploring this capability more fully. Finally, as discussed in the results section, a receding horizon formulation of \algo would help reduce goal-errors and better support navigation with exploration. We leave this as future work.


\section{Conclusion}\label{sec:conc}

We formulate the problem of simultaneous trajectory estimation and planning (\algo) as probabilistic inference. By representing the prior distribution of a continuous-time trajectory and likelihood function of costs and observations with factor graphs, we can efficiently perform inference to compute the posterior distribution of the trajectory. We solve STEAP in an online setting to simultaneously estimate and smooth the trajectory history as well as replan for the future trajectory as new information is encountered. This is made possible by performing efficient incremental inference to update the previous solution. We conducted experiments in simulation and on a real mobile manipulator and showed that our framework is able to perform in real-time and robustly handle the stochasticity associated with execution. Our results demonstrate that this framework is suitable for online applications with high-degree-of-freedom systems in known, static real-world environments.


\begin{acknowledgements}
This work was partially supported by NSF NRI award 1637908 and National Institute of Food and Agriculture, USDA, award 2014-67021-22556. The authors thank Muhammad Asif Rana, David Kent, Vivian Chu, and Sonia Chernova for access to and help with the Vector robot.
\end{acknowledgements}


\bibliographystyle{spbasic}
\bibliography{ref}


\end{document}